\documentclass{elsarticle}
\usepackage[vmargin=2cm]{geometry}
\usepackage{multirow}
\usepackage{float}
\floatstyle{plaintop}
\restylefloat{table}
\restylefloat{figure}
\usepackage{multirow}
\usepackage{pdflscape}
\usepackage{xcolor}

\usepackage{graphicx}
\usepackage{caption}
\usepackage{subcaption}
\usepackage{amssymb}
\usepackage{amsmath,}
\usepackage{physics}
\usepackage{algorithm}
\usepackage{algorithmic}
\usepackage{longtable}
\usepackage{lineno}

\journal{Neural Networks, Elsevier}

\begin{document}

\begin{frontmatter}


\title{Oblique and rotation double random forest }



 \author[label1]{M.A. Ganaie}
 \ead{phd1901141006@iiti.ac.in}
\author[label1]{M. Tanveer\corref{cor1}}
\ead{mtanveer@iiti.ac.in}
\cortext[cor1]{Corresponding author}
\address[label1]{Department of Mathematics, Indian Institute of Technology Indore, Simrol, Indore, 453552, India}
\address[label2]{School of Electrical \& Electronic Engineering, Nanyang Technological University, Singapore} 
\address[label4]{KINDI Center for Computing Research, College of Engineering, Qatar University, Qatar\fnref{label4}}
\address[label3]{Department of Computer Science, VŠB - Technical University of Ostrava, Czech Republic}




\author[label2,label4]{P.N. Suganthan\corref{cor1}}
\ead{epnsugan@ntu.edu.sg}

\author[label3]{V. Snasel}
\ead{vaclav.snasel@vsb.cz}

\begin{abstract}
Random Forest is an ensemble of decision trees based on the bagging and random subspace concepts. As suggested by Breiman, the strength of unstable learners and the diversity among them are the ensemble models' core strength. In this paper, we propose two approaches known as oblique and rotation double random forests. In the first approach, we propose rotation based  double random forest. In rotation based double random forests,  transformation or rotation of  the feature space is generated at each node. At each node different random feature subspace is chosen for evaluation, hence the transformation at each node is different. Different transformations result in better diversity among the base learners and hence, better generalization performance. With the double random forest as base learner,  the data at each node is transformed via two different transformations namely, principal component analysis and linear discriminant analysis. 
In the second approach, we propose oblique double random forest.  Decision trees in random forest and double random forest are univariate, and this results in the generation of axis parallel split which fails to capture the geometric structure of the data. Also, the standard random forest may not grow sufficiently large decision trees resulting in suboptimal performance. To capture the geometric properties and to grow the decision trees of sufficient depth, we propose oblique double random forest. The  oblique double random forest  models are multivariate decision trees. At each non-leaf node, multisurface proximal support vector machine generates the optimal plane for better generalization performance. Also, different regularization techniques (Tikhonov regularisation, axis-parallel split regularisation, Null space regularisation) are employed for tackling the small sample size problems in the decision trees of  oblique double random forest. 
The proposed ensembles of decision trees produce trees with bigger size compared to the standard ensembles of decision trees as bagging is used at eah which results in improved performance.
The evaluation of the baseline models and the proposed oblique and rotation double random forest models is performed on benchmark $121$ UCI datasets and real-world fisheries datasets. Both statistical analysis and the experimental results demonstrate the efficacy of the proposed oblique and rotation double random forest models compared to the baseline models on the benchmark datasets.
\end{abstract}

\begin{keyword}
Double random forest \sep Oblique random forest \sep Support vector machine \sep Bias \sep Ensemble \sep Oblique \sep Orthogonal \sep Classification \sep Classifiers \sep Ensemble learning \sep Random forest \sep Bootstrap \sep Decision tree.


\end{keyword}

\end{frontmatter}


\section{Introduction}
\label{S:1}
Perturb and combine approach \cite{breiman1996bias} is the core of the ensemble strategy \cite{dietterich2000ensemble} and hence, it has been used across different domains like machine learning \cite{wiering2008ensemble}, computer vision tasks \cite{goerss2000tropical} for recognition of patterns,  mining big data \cite{lulli2019mining} and biomedical domain \cite{pal2021prediction}. Both theoretical and empirical aspects of the ensemble learning have been explored in the literature. Multiple classifier systems \cite{zhou2013multiple} or ensemble learning perturbs the input data to induce diversity among the base learners of an  ensemble and uses combine strategy to aggregate the outputs of base learners such that the generalization  of the ensemble model is superior in comparison with the individual learners. 

To analyze how the ensemble learning performs better compared to individual models,  studies  like  reduction in variance among the base learners \cite{breiman1996bias,geurts2006extremely,zhang2008rotboost} have been putforth. With the bias and variance reduction theory \cite{breiman1996bias,kohavi1996bias}, the error in classification  is given in terms of bias and variance. Bias measure gives how far is the average guess of each base learner from the target class over the perturbed training sets generated from a given training set and variance measure is how much the base learners  guess fluctuates with the perturbations of the given training set.

Decision tree algorithm is a commonly used classification model due to its simplicity and better interpretability.  Decision tree uses divide and conquer approach to recursively partition the data.  The recursive partition of the tree is sensitive to perturbation of the input data, and results in an unstable classifier. Hence, it is said to have high variance and low bias. The ensemble methodology can be used in unstable classifiers to further improve the classification performance.  

Random forest \cite{breiman2001random} and rotation forest \cite{rodriguez2006rotation} are the well-known classification models, widely used in the literature. Both these models are based on the ensemble methodology and use decision tree as the base classifier. Due to the better generalization performance, random forest proved to be one of the best classification models among $179$ classifiers evaluated on  $121$ datasets \cite{fernandez2014we}. 

 An ensemble of decision trees, Random forest,  uses bagging \cite{breiman1996bagging} and random subspace \cite{ho1998random} strategy. These two approaches induce diversity among the base learners, here decision trees, for better generalization. Bagging, also known as bootstrap aggregation, generates multiple bags of a given training set such that each decision tree is trained on a given bag of the data. Each tree uses a bag of training data whose distribution is akin to the whole population and hence, each classifier shows good generalization performance. Within each decision tree, random subspace approach is used in each non-terminal node to further boost the diversity among the base models. Random forest has been successfully applied for analysis of gene expression data \cite{jiang2004joint}, EEG classification \cite{shen2007feature}, spectral data classification \cite{menze2009comparison}, recognition of objects, image segmentation \cite{ho1998random,hothorn2005design} and chimera identification \cite{ganaie2020identification}. Other applications include selection of features \cite{menze2009comparison}, analysis of sample proximities \cite{menze2007multivariate} and so on. Random forest have also been adapted to Spark based distributed and scalable environments \cite{lulli2017crack,lulli2017reforest}. With the growing of privacy concerns, Random forest models have been improved to meet  the  privacy expectations. Differential privacy \cite{dwork2008differential} has been widely adopted in Random  Forest \cite{patil2014differential,fletcher2017differentially,guan2020differentially,xin2019differentially}.

To obtain the better generalization performance, various hyperparameters of the random forest need to be chosen optimally. These hyperparameters include number of base learners (here, decision trees) in a forest (ntree), number of candidate features for evaluation at a given non-leaf node (mtry), and number of samples in an impure node (nodesize or minleaf) (we will use minleaf and nodesize interchangeably). To get these parameters optimally, different studies have been proposed. Analysis of tuning process \cite{probst2017tune, freeman2016random},  sensitivity of the parameters \cite{ huang2016parameter}, effect of number of trees in an ensemble \cite{banfield2006comparison,hernandez2013large, oshiro2012many} provide insight how these parameters affect the model performance. To obtain the optimal number of candidate features, different methods  \cite{boulesteix2012overview,han2019optimal} have been proposed. Analysis of optimal sample size in bagging \cite{martinez2010out} and estimation of tree size via combination of random forest with adaptive nearest neighbours \cite{lin2006random} result in the better choice of the hyperparameters. 

Broadly speaking, there are two approaches, namely, univariate decision trees \cite{banfield2006comparison} and multivariate decision trees \cite{murthy1995growing} for generating the decision trees. Univariate decision trees, also known as axis parallel or orthogonal decision trees, use some impurity criteria to optimize best univariate split feature among the set of randomly chosen subspace of features. Multivariate decision trees, also known as oblique decision trees, perform the node splitting using all or a part of the features. In general, decision boundary of an oblique decision tree can be approximated by a large number of stair-like decision boundaries of the  univariate decision trees.

Random forest is a univariate model and builds hyperplane at each non-terminal node  such that splitting at the children nodes is easier in a given decision tree.  At a given non-leaf node, the splitting hyperplane  may not be a good classifier \cite{manwani2011geometric}. Different criteria like entropy measure, Gini index measure  and twoing rule are involved in most of the decision tree based models for choosing the best split among the set of splits such that the best split results in lowest impurity score.
 At each non-leaf node, impurity criteria  measures skewness of the distribution of the different category samples. Nearly uniform distribution is assigned low impurity score while as high impurity score is given to a distribution wherein the samples of a particular class dominate the other classes. In most of the decision tree based induction tree algorithms, some impurity measure is optimized for generating the tree. However, due to non differentiability of the impurity measures with respect to the hyperplane parameters, different search techniques are employed for generating the decision trees. Like deterministic hill-climbing model in CART-LC \cite{breiman1984classification}, randomized search based CART-LC in OC1 \cite{murthy1993oc1}. In high dimensional feature space, both these methods suffer due to searching in 1-D at a time and local optimum problem. Thus, to avoid the local optima, multiple trails or restarts are used to minimize the chances of ending up with the local optima. Evolutionary approaches have also been used for optimizing in all dimensions \cite{pedrycz2005genetically,cha2009genetic} which  tolerate  the noisy evaluation of a rating function and also simultaneously optimize  the multiple rating functions \cite{cantu2003inducing,pangilinan2011pareto}. Extremely randomized trees \cite{geurts2006extremely} and its oblique version \cite{zhang2014towards}, strongly randomized  the attribute set and its cut point. Other approaches include fuzzy based decision trees \cite{wang2008induction,wang2008improving}, ensemble of feature spaces \cite{zhang2014random} and  decision tree support vector machine \cite{zhang2007decision}. 
  Random feature weights for decision tree ensemble \cite{maudes2012random} associates weight to each attribute for better diversity of the model. Recent studies have evaluated the interpretability of the decision forests so that the decisions can be interpreted for better understanding \cite{sagi2020explainable,fernandez2020random}.
 For more literature about the decision trees, we refer the readers to \cite{rokach2016decision}.

With all the impurity measures, given in \cite{manwani2011geometric}, the issue is  they are function of  different class distributions on each side of the hyperplane and ignore the geometric structure of the class regions \cite{manwani2011geometric} as if the impurity measure is unaffected if one changes the data labels without any change in the relevant features of each category on either side of the hyperplane.  

To incorporate the geometric structure of class distributions, support vector machines (SVM) \cite{cortes1995support} are employed to generate the decision trees \cite{manwani2011geometric}. The multisurface proximal support vector machines (MPSVM) \cite{mangasarian2005multisurface} generate the proximal hyperplanes in a manner that each plane  is proximal to the samples of one class and farthest  from the samples of another class. \citet{manwani2011geometric} generated the two clustering planes at each non-leaf node and choose the angle bisector of these planes which makes the nodes pure. MPSVM is a binary class algorithm, hence, they decomposed the multiclass problem into a binary class  by grouping the majority class samples into one class and rest samples into other class. As the node becomes purer with the growth of a tree, the subsequent nodes receive smaller number of samples. To avoid this problem, NULL space method \cite{chen2000new} is used in \cite{manwani2011geometric}. Also, MPSVM based oblique decision tree ensemble \cite{zhang2014oblique} employed regularisation approaches like Tikhonov
regularization \cite{marroquin1987probabilistic} and axis-parallel split regularization. In \cite{ganaie2020oblique}, twin bounded SVM \cite{shao2011improvements} resulted in more generalization performance as no explicit regularisation methods are needed to handle these problems. Both MPSVM based oblique decision tree ensemble \cite{zhang2014oblique}  and  TBSVM based oblique decision tree ensemble \cite{zhang2014oblique} use single base learner at each nonleaf node to search the optimal split among the candidate splits. The oblique decision tree ensemble showed better generalization than the standard random forest \cite{zhang2017benchmarking}. Heterogeneous oblique random forest \cite{katuwal2020heterogeneous} generates hyperplanes via MPSVM, logistic regression, linear discriminant analysis, least squares SVM and ridge regression. The optimal hyperplane for best split is chosen among the generated planes which results in purer nodes. 

Recent study of double random forest \cite{han2020double} evaluated the effect of node size on the performance of the model. The study revealed that the prediction performance may improve if  deeper decision trees are generated. The authors showed that the largest tree grown on a given data by the standard random forest might not be sufficiently large to give the optimal performance. Hence, double random forest \cite{han2020double} generated decision trees that are bigger than the ones in standard random forest. The maximum performance of the random forest is achieved corresponding to the minimum node size which generates the larger trees \cite{zhang2014oblique}. This supports the hypothesis that  larger the trees of an ensemble the better the performance of the model is. 
Instead of training each decision tree with different bags of training set obtained via bagging approach at the root node,  \citet{han2020double} generated each tree with the original training set and used bootstrap aggregation at each non-terminal node of the decision tree to obtain the best split. However, both the random forest and double random forest are univariate decision trees and hence ignore the geometric class distributions resulting in lower generalization performance. To overcome these issues, we propose  oblique  double random forest.   Oblique double random forest  models integrate the benefits of double random forest and the geometric structure information of the class distribution for better generalization performance. 
For generating more diverse ensemble learners  in the double random forest,  feature space is rotated or transformed at each non leaf node using two transformations known as linear discriminant analysis and principal component analysis. Using transformations at each non-leaf node on different randomly chosen feature subspaces improves diversity among the base models and  leads to better generalization performance.  

The main highlights of this paper are:
\begin{itemize}
    \item We use different rotations (principal component analysis and linear discriminant analysis) at each non-leaf node to generate diverse double random forest ensembles (DRaF-PCA and DRaF-LDA).

    \item The proposed oblique double random forest (MPDRaF-T, MPDRaF-P and MPDRaF-N) variants use MPSVM for obtaining the optimal separating hyperplanes at each non-terminal node of the decision tree ensembles.
    
    
    \item The proposed ensemble of double Random forest generate larger trees compared to the variants of standard Random forest. 
    
    \item Statistical analysis reveals that the average rank of the proposed double random forest models is superior than the standard random forest.  Moreover, the average accuracy of the proposed DRaF-LDA, DRaF-PCA, and MPDRaF-P is superior than the standard random forest and standard double random forest models. Also, the average rank of the proposed DRaF-LDA, MPDRaF-P and DRaF-PCA is better compared to the standard double Random forest.
\end{itemize}

\section{Related work}
\label{Sec:Related work}
In this section, we briefly review the related work of the ensemble of decision trees. 

\begin{table}[]
    \centering
    \resizebox{\textwidth}{!}{
    \begin{tabular}{cl}
    \hline
    ABBREVIATION &DEFINITION\\
    \hline
    PCA & Principal component analysis \\
  LDA&  Linear discriminant analysis \\
  SVM & Support vector machines\\
         RaF &Standard Random Forest\\
DRaF &Standard double Random Forest\\
MPSVM &Multisurface proximal support vector machines\\
RaF-PCA& Principal component analysis based ensemble of decision trees\\ 
RaF-LDA &Linear discriminant  analysis based ensemble of decision trees \\
MPRaF-T &MPSVM based oblique decision tree ensemble with Tikhonov regularisation\\
MPRaF-P &MPSVM based oblique decision tree ensemble with axis parallel regularisation\\
MPRaF-N &MPSVM based oblique decision tree ensemble with NULL space regularisation\\
DRaF-PCA &Rotation based double random forest with principal component analysis \\
DRaF-LDA &Rotation based double  random forest with linear discriminant analysis \\
MPDRaF-T & Oblique double random forest  with MPSVM via Tikhonov regularisation\\
MPDRaF-P & Oblique double random forest  with MPSVM via axis parallel regularisation\\
MPDRaF-N & Oblique double random forest  with MPSVM via NULL space regularisation\\
\hline
    \end{tabular}}
    \caption{Nomenclature}
    \label{tab:Nomenclature}
\end{table}

\subsection{Handling multiclass problems}
MPSVM is a binary classification model and finding  the optimal separating hyperplanes at each non-terminal node of a decision tree may be a multiclass problem. To handle the multiclass problem via binary class approach, different methods like one-versus-all \cite{bottou1994comparison}, one-versus-one \cite{knerr1990single}, decision directed acyclic graph \cite{platt1999large}, error correcting output codes \cite{dietterich1994solving} and so on have been proposed.  Data partitioning rule of the decision trees at each non-leaf node proves handy over other binary classification models \cite{zhang2014oblique}. Separating the  classes  with majority samples as one class and rest samples as another class results in an inefficient model as it fails to capture the geometric structure of the data samples \cite{manwani2011geometric}. To incorporate the geometric structure, the authors in \cite{zhang2014oblique} decomposed the multiclass problem into a binary one by using  class separability information. 
The authors used Bhattacharyya distance for decomposition. In statistics, Bhattacharyya distance gives the measure of similarity between the two  discrete probability distributions or continuous probability distributions  as it is deemed to be a good insight about separability of classes between two normal classes $C_1 \sim N(\mu_1,\nu_1),C_2 \sim N(\mu_2,\nu_2)$, where $\mu_i$ and $\nu_i$ are the parameters of the normal distribution of class $C_i$, for $i=1,2$. Following the similar approach as in \cite{zhang2014oblique}, we used multivariate Gaussian distribution \cite{jiang2011linear}. Motivated by \cite{zhang2014oblique,jiang2011linear}, we use Bhattacharyya distance to measure the class separability for decomposing the multiclass problem into a binary class problem (Algorithm \ref{algo:Multiclass to binary class}). 

\begin{algorithm}
\caption{Decomposition of Multiclass problem  to a binary class problem}
\label{algo:Multiclass to binary class}
\textbf{Input:}

     $D:=N\times n$ be the training dataset with $N$ number of data points with feature size $n$. \\
$Y:=N\times 1$ be the target labels. \\
$\{L_1, L_2, \dots, L_C\}$ be the target labels.

\textbf{Output:}\\
$C_p$ and $C_n$ are two hyperclasses or groups
\begin{algorithmic} 
\STATE For each class $j=1,2,\dots, C$.
\begin{enumerate}
    \item For each pair of $L_j$ and $L_k$, for $k=j+1,\dots,C$ as:
    \begin{align}
\label{eqn:Bhattacharyya}
    F(L_j,L_k)=\frac{1}{8}(\mu_k-\mu_j)^t\Big(\frac{\nu_j+\nu_k}{2}\Big)^{-1}(\mu_k-\mu_j)+\frac{1}{2}ln \frac{|(\nu_j+\nu_k)/2|}{\sqrt{|\nu_j||\nu_k|}}
\end{align}
  \item Find the pair $L_p$ and $L_n$ of classes with the maximum Bhattacharyya distance, and assign them to $C_p$ and $C_n$ respectively.
  \item For every other class, if $F(L_k,L_p)<F(L_k,L_n)$ then group $L_k$ to $C_p$ otherwise  group in $C_n$.
\end{enumerate}
\end{algorithmic}
\end{algorithm}

\subsection{Multisurface proximal support vector machine}
Multisurface proximal support vector machine (MPSVM) \cite{mangasarian2005multisurface} is a binary class algorithm. Suppose $X_1, X_2$ be the data points belonging to the positive and negative class, respectively. Here, $X_1 \in \mathbb{R}^{m_1\times n}$, $X_2 \in \mathbb{R}^{m_2\times n}$ and each sample $x\in \mathbb{R}^n$.  MPSVM generates two hyperplanes as 
\begin{align}
\label{eqn:MPSVM2}
    x^tw_1-b_1=0 ~~ \text{and}~~x^tw_2-b_2=0,
\end{align}
where $(w_1,b_1)$ and $(w_2,b_2)$ are the planes closer to the samples of positive and negative class, respectively. MPSVM minimises the sum of squared two norm distances between the samples of positive class divided by the sum of squared distances from the samples of negative class to the plane. Thus, the optimization problems of MPSVM are given as follows:
\begin{align}
\label{eqn:MPSVM3}
    \underset{(w,b)\neq 0}{min}~~\frac{\norm{X_1w-eb}^2/\norm{\begin{matrix}w\\b\end{matrix}}^2}{\norm{X_2w-eb}^2/\norm{\begin{matrix}w\\b\end{matrix}}^2} 
\end{align}
and
\begin{align}
\label{eqn:MPSVM4}
    \underset{(w,b)\neq 0}{min}~~\frac{\norm{X_2w-eb}^2/\norm{\begin{matrix}w\\b\end{matrix}}^2}{\norm{X_1w-eb}^2/\norm{\begin{matrix}w\\b\end{matrix}}^2}, 
\end{align}
where $\norm{\cdot}$ is a two norm, $e$ is a vector of ones with appropriate dimensions.

Suppose 
\begin{align}
\label{eqn:MPSVM5}
    P=\begin{bmatrix}A&-e \end{bmatrix}^t\begin{bmatrix} A&-e\end{bmatrix}, Q=\begin{bmatrix} B &-e\end{bmatrix}^t \begin{bmatrix}B&-e \end{bmatrix}, r=\begin{bmatrix} w\\b\end{bmatrix},
\end{align} then the optimization problem \eqref{eqn:MPSVM3} is given as
\begin{align}
\label{eqn:MPSVM6}
    \underset{r\neq0}{min}~~\frac{r^tPr}{r^tQr}.
\end{align}
Similarly, the optimization problem \eqref{eqn:MPSVM4} is given as follows:
\begin{align}
\label{eqn:MPSVM8}
    \underset{r\neq0}{min}~~\frac{r^tSr}{r^tUr},
\end{align}
where $S=\begin{bmatrix}B&-e \end{bmatrix}^t\begin{bmatrix}B&-e \end{bmatrix}$ and $U=\begin{bmatrix}A&-e \end{bmatrix}^t\begin{bmatrix}A&-e \end{bmatrix}$.

The clustering hyperplanes are obtained by solving the following generalized eigenvalue problems:
\begin{align}
\label{eqn:MPSVM9}
    Pr&=\lambda Qr, ~r\neq 0\\
    \label{eqn:MPSVM9a}
    Sr&=\gamma Ur, ~r\neq0.
\end{align}
The optimal hyperplanes are the eigenvectors corresponding to the smallest eigenvalues.

The way \eqref{eqn:MPSVM9} and \eqref{eqn:MPSVM9a} are defined, the clustering hyperplanes are able to capture the geometric properties of the data which  are helpful while discriminating among the classes.

\subsubsection{Random Forest}
Random forest \cite{breiman2001random} is an ensemble with decision tree as the base learner which are generated using the concept of bagging and random subspace method. Both bagging and random subspace methods induce diversity among the decision trees of an ensemble. Each decision tree of an ensemble chooses the optimal split among the  randomly selected candidate feature subsets at a given non-leaf node. The optimal split is chosen using some impurity criterion's  like information gain, Gini impurity and so on \cite{breiman1984classification}.

The algorithm of the random forest is given in Algorithm \ref{algo:Random Forest}. The classification and regression tree (CART) \cite{breiman2001random} performs the test split using only one feature and hence, known as univariate decision tree \cite{murthy1995growing}.

\begin{algorithm}
\caption{Random Forest}
\label{algo:Random Forest}
\textbf{Training Phase:}\\
\textbf{Given:}
\begin{itemize}
    \item []
 $D:=N\times n$ be the training dataset with $N$ number of data points with feature size $n$. \\
$Y:=N\times 1$ be the target labels. \\
$L:$ is number of base learners. \\
\textit{``mtry"}: number of candidate features to be evaluated at each non-leaf node. \\
\textit{``nodesize"} or \textit{``minleaf''}: maximum number of samples in an impure node.
\end{itemize}
\begin{algorithmic} 
\STATE For each decision tree, $T_i$ for $i=1,2,\dots, L$
\begin{enumerate}
    \item Generate bootstrap samples $D_i$ from D.
    \item Generate the decision tree using $D_i$:\\
         For a given node $d$:
         \begin{enumerate}
             \item[(i)] Choose \textit{``mtry"}$=\sqrt{n}$ number of features from the given feature space of $D_i$.
             \item[(ii)] Select the best feature split feature and the cutpoint among the random feature subset.
             \item[(iii)] With the optimal split feature and the cutpoint, divide the data.
         \end{enumerate}
         Repeat steps (i)-(iii), until the stopping criteria is met.
\end{enumerate}

\end{algorithmic}
\textbf{Classification Phase:}\\
For a test data point $x_i$, use the base learner of the forest to generate the label of the test sample. The predicted label of the test data point is given by the majority voting of the decision trees of an ensemble.
\end{algorithm}

\begin{algorithm}
\caption{ Double Random Forest}
\label{algo:Double Random Forest}
\textbf{Training Phase:}\\
\textbf{Given:}\\
$D:=N\times n$ be the training dataset with $N$ number of data points with feature size $n$. \\
$D_i:=N_i\times n_i$ be the training samples reaching to a node $i$, with $N_i$ number of samples with feature size $n$. \\
$Y:=N\times 1$ be the target labels. \\
$L:$ is number of base learners. \\
\textit{``mtry"}: number of candidate features to be evaluated at each non-leaf node. \\
\textit{``nodesize"} or \textit{``minleaf''}: maximum number of data samples to be placed in an impure node.
\begin{algorithmic} 
\STATE For each decision tree, $T_i$ for $i=1,2,\dots, L$
\begin{enumerate}
    \item Use training data $D$.
    \item Generate the decision tree $T_i$ with randomly chosen  subset of features and randomised bootstrap instance using $D$:\\
         For a given node $d$ with data $D_d$:
         \begin{enumerate}
             \item[(i)]
             if $N_d>N \times 0.1$ \\
             $~~$Generate bootstrap sample $D_d^*$ from $D_d$.\\
                  else \\
                  $~~$$D_d^*=D_d$
                  
             \item[(i)] Choose \textit{``mtry"}$=\sqrt{n}$ number of features from the given feature space of $D_d^*$. 
             
             \item[(ii)]
             Select the best split feature and the cutpoint among the random feature subset $D_d^*$.
             \item[(iii)] With  the optimal split feature and the cutpoint with $D_d^*$, split the data $D_d$ into child nodes.
         \end{enumerate}
         Repeat steps (i)-(iii), until either of the satisfied: \\
         \begin{itemize}
             \item Node reaches to  purest form.
             \item Samples reaching a given node are lesser or equal than \textit{minleaf}
         \end{itemize}
         
\end{enumerate}

\end{algorithmic}
\textbf{Classification Phase:}\\
For a test data point  $x_i$, use the decision trees of the forest to generate the label of the test sample. The predicted class of the test data point is given by the majority voting of the decision trees of an ensemble. 
\end{algorithm}

\subsection{Double random forest}
Double random forest \cite{han2020double} is an ensemble with decision tree as the base learner which uses the concept of bagging and the random subspace method. Unlike standard random forest wherein the base learner is trained on the boostrapped sample of the dataset, double random forest trains each base learner on  the original dataset. This results in more unique features in the data used in training the double random forest than standard forest. The more number of unique instances leads to larger decision trees and hence better generalization performance.
Double random forest uses bootstrap sampling momentarily at every non-terminal node. Once the feature which gives the split is chosen among the randomly chosen subset of the features from the bootstrap samples, the splitting of the original data is done and hence original data is sent down the decision tree resulting in more number of unique instances. The algorithm of the double random forest is given in Algorithm \ref{algo:Double Random Forest}.

\begin{algorithm}
\caption{ Null Space Regularization}
\label{algo:Null Space Regularization}

\textbf{Input:}
$P$ (Positive class) and $H$ (Negative class) as given in \eqref{eqn:MPSVM5}.\\
\textbf{Output:} Clustering hyperplane $\begin{bmatrix}w\\b \end{bmatrix}$.
\begin{enumerate}
    \item Suppose $P$ is rank deficit with rank $r<n+1$, calculate $O=[\alpha_1, \alpha_2, \cdots,\alpha_{n+1-r}]$  whose columns are the orthonormal basis for the Null space of $P$. 
    \item Project the matrix $Q$ in the Null space of $P$. For each vector (row) $p$ in matrix $P$, the projection is given as $pOO^t$. Hence, the projection of matrix $Q$ is given as $\Bar{Q}=\sum_{p\in Q} OO^tp^tpOO^t=OO^tQOO^t$. In the similar manner, the projeection of matrix $P$ is given as $\Bar{P}=OO^tPOO^t$.
    \item Since the columns of $O$ span the Null space of $P$, hence $\Bar{P}$ would be zero. Thus, the desired plane is the eigen vector corresponding to the largest eigenvector of $\Bar{Q}$.
\end{enumerate}

\end{algorithm}

\section{Proposed oblique and rotation double random forest}
\label{Sec:Proposed Method}
This paper  proposes two approaches for generating the oblique and rotation double random forest known as  oblique  double random forest models and the rotation based  double random forest models. Two approaches are given as follows:
 
\subsection{ Oblique double random forest  with MPSVM}
Univariate decision trees don't capture properties of the data geometrically. Both standard random forest and double random forest are univariate decision tree ensembles. Also, decision trees in the standard random forest may not be large enough for the  datasets to get the better generalization. To overcome these limitations, we propose  oblique double random forest  with MPSVM. Unlike standard random forest, the oblique double random forest models with MPSVM use bootstrapping samples at every  non-terminal node (until some condition is met as given in Algorithm \ref{algo:Oblique Double Random Forest}) for generating the optimal oblique splits and divide the original data instead of bootstrapped samples among the children nodes. 
 To incorporate the geometric structure in the splitting hyperplane, the proposed oblique double random forest uses MPSVM wherein optimal split at each non-leaf node is generated based on the clustering hyperplanes. As the decision tree size increases, the  data points arriving at a particular node decreases and hence, the issues of sample size may arise. To overcome this issue, we use different regularization techniques to obtain a better generalization performance. The  regularization approaches used are Tikhonov regularization, axis parallel split regularization and null space approach. If the model uses Tikhonov regularization then the proposed model is named as oblique double random forest via MPSVM with Tikhonov regularization (MPDRaF-T), if the model uses axis parallel split regularization then the proposed model is known as oblique double random forest via MPSVM with  axis parallel split regularization (MPDRaF-P) and if the model uses null space approach then the proposed models is known as oblique double random forest via MPSVM with  null space approach (MPDRaF-N).
 In Tikhonov regularization, the small positive number is added along the diagonal elements to regularize the data matrix (say, $H$) i.e., if data matrix $H$ is rank deficient, then regularize $H$ as :
\begin{align}
    H=H+\delta \times I,
\end{align}
where $\delta$ is a small positive number and $I$ is appropriate dimensional identity matrix.
In axis-parallel split regularization, if the data matrix (say, $H$) is rank deficient at a given node then we follow axis parallel approach to complete the growth of decision tree. Thus, heterogeneous test functions are used for growing the decision trees. i.e., till the current node MPSVM is used for generating the optimal splits and now onwards axis parallel approach is followed for growing the decision tree.
In order to handle the sampling issues, \citet{manwani2011geometric} proposed the Null space approach (given in Algorithm \ref{algo:Null Space Regularization}) for regularizing the matrices. For the proposed MPDRaF-N, we follow the Algorithm \ref{algo:Null Space Regularization} for regularizing the matrices.

 Algorithm \ref{algo:Oblique Double Random Forest} summarises the  oblique double random forest  with MPSVM.

\begin{algorithm}
\caption{Oblique  Double Random Forest with MPSVM}
\label{algo:Oblique Double Random Forest}
\textbf{Training Phase:}\\
\textbf{Given:}

$D:=N\times n$ be the training set with $N$ number of samples with feature size $n$. \\
$D_i:=N_i\times n_i$ be the training samples reaching to a node $i$, with $N_i$ number of samples with feature size $n_i$. \\
$Y:=N\times 1$ be the target labels. \\
$L:$ is number of base learners. \\
\textit{``mtry"}: number of candidate features to be evaluated at each non-leaf node. \\
\textit{``nodesize"} or \textit{``minleaf''}: maximum number of data samples to be placed in an impure node.
\begin{algorithmic} 
\STATE For each decision tree, $T_i$ for $i=1,2,\dots, L$
\begin{enumerate}
    \item Use training data $D$.
    \item Generate the decision tree $T_i$ with randomly chosen  subset of features and randomised bootstrap instance using $D$:\\
         For a given node $d$ with data $D_d$:
         \begin{enumerate}
             \item[(i)]
             if $N_d> N \times 0.1$ \\
             $~~$Generate bootstrap sample $D_d^*$ from $D_d$.\\
                  else \\
                  $~~$$D_d^*=D_d$
                  
             \item[(i)] Choose \textit{``mtry"}$=\sqrt{n}$ number of features from the given feature space of $D_d^*$ 
             \item[(ii)] Using Algorithm \ref{algo:Multiclass to binary class} group the dataset $D_d^*$ into  $C_p$ and $C_n$.
             \item[(iii)] Use MPSVM (with different regularization's) for generating  the optimal split with $C_p$ and $C_n$ as input, and split the data $D_d$ into child nodes.
         \end{enumerate}
         Repeat steps (i)-(iii), until the stopping criteria is one of the conditions is met: \\
         \begin{itemize}
             \item Node reaches to  purest form.
             \item Samples reaching a given node are lesser or equal than \textit{minleaf}
         \end{itemize}
\end{enumerate}
\end{algorithmic}
\textbf{Classification Phase:}\\
For a test data point  $x_i$, use the decision trees of the forest to generate the label of the test sample. The predicted class of the test data point is given by the majority voting of the decision trees of an ensemble. 
\end{algorithm}

\subsection{Double random forest with PCA/LDA}
For generating the diverse learners in an ensemble, we propose rotation based double random forest ensemble models.
Rotation or transformation on different random feature subspaces results in different projections leading to better generalization performance.
In this method, the objective is to rotate or transform the data for better diversity among the base learners. At each non-leaf node, the rotation is applied on random feature subspace which results in improved diversity among the base classifiers. We use two approaches for rotation of feature subspace i.e., principal component analysis (PCA) and linear discriminant analysis (LDA). 

The proposed double random forest with PCA (DRaF-PCA) is given in Algorithm \ref{algo:Double Random Forest_PCA}.  At each non-leaf node, rotation or transformation is applied on the bootstrapped samples reaching a given node with random feature subspace.

The algorithm of the proposed double random forest with LDA (DRaF-LDA) varies from Algorithm \ref{algo:Double Random Forest_PCA} at step $(ii)$ and $(iii)$. In DRaF-LDA model, instead of calculating total scatter matrix $S_d$ at each node, within class scatter matrix $S_d^w$ and between class scatter matrix $S_d^b$ are calculated. Then, generalized eigenvectors of $(S_d^w,S_d^b)$ are calculated ($S_d^b \times \alpha =\lambda\times S^w_d$, where $\alpha$ is the generalized eigenvector corresponding to the generalized eigenvalue $\lambda$).

\begin{algorithm}
\caption{Double Random Forest with PCA}
\label{algo:Double Random Forest_PCA}
\textbf{Training Phase:}\\
\textbf{Given:}

$D:=N\times n$ be the training set with $N$ number of samples with feature size $n$. \\
$D_i:=N_i\times n_i$ be the training samples reaching to a node $i$, with $N_i$ number of samples with feature size $n_i$. \\
$Y:=N\times 1$ be the target labels. \\
$L:$ is number of base learners. \\
\textit{``mtry"}: number of candidate features to be evaluated at each non-leaf node. \\
\textit{``nodesize"} or \textit{``minleaf''}: maximum number of data samples to be placed in an impure node.

\begin{algorithmic} 
\STATE For each decision tree, $T_i$ for $i=1,2,\dots, L$
\begin{enumerate}
    \item Use training data $D$.
     \item Generate the decision tree $T_i$ with randomly chosen  subset of features and randomised bootstrap instance using $D$:\\
         For a given node $d$ with data $D_d$:
         \begin{enumerate}
             \item[(i)]
             if $N_d>N \times 0.1$ \\
             $~~$Generate bootstrap sample $D_d^*$ from $D_d$.\\
                  else \\
                  $~~$$D_d^*=D_d$
                  
             \item[(i)] Choose \textit{``mtry"}$=\sqrt{n}$ number of features from the given feature space of $D_d^*$ 
             \item[(ii)] Calculate total scatter matrix $S_d$ using $D_d^*$.
             
             \item[(iii)] Calculate all the eigenvectors of $S_d$, denoted by $V$.
             
             \item[(iv)] Calculate the data transformation using all the eigenvectors $V$ as, $D^*_{PCA}=D^*_d*V.$

             \item[(v)]
             In the PCA space, search the best feature split.
             \item[(iii)] With the optimal split feature and the cutpoint, split the data $D_d$ into the child nodes.
         \end{enumerate}
         Repeat steps (i)-(iii), until the stopping criteria is met.
\end{enumerate}

\end{algorithmic}
\textbf{Classification Phase:}\\
For a test sample $x_i$, generate labels via decision trees of the forest. 
At every non-terminal node, the test data sample is rotated with the same matrix $V$ generated in the training stage.
The predicted class of the test data point is given by the majority voting of  decision trees of an ensemble.
\end{algorithm}

\section{Comparison of the proposed oblique and rotation based double random forest models with the existing baseline models}
The main differences of the proposed models with respect to the existing models are given as follows: 
\begin{enumerate}
    \item MPDRaF-T, P, N are the oblique double random forest variants which employ bagging at each non leaf node to allow the generation of bigger trees. Unlike standard variants like RaF, MPRaF-T, MPRaF-P and MPRaF-N, the proposed models use the training bags which have more unique instances of the samples which results in generation of bigger trees. Moreover, MPDRaF-T,P,N capture the geometric properties of the data which is ignored by the standard RaF and double RaF models.
    \item The standard RaF and DRaF models use the concepts of random subspace and bagging for introducing the diversity among the base learners of an ensemble. However, the proposed DRaF-PCA and DRaF-LDA employ PCA and LDA transformations at non-leaf nodes in addition to the random subspace and bagging  concepts for producing more diverse base learners. Thus, the proposed DRaF-PCA and DRaF-LDA models possess better diversity compared to the RaF and DRaF models. Unlike RaF-PCA and RaF-LDA, the proposed DRaF-PCA and DRaF-LDA models use bagging concept at each non-leaf node which allow greater depth of the tree and hence better performance.
\end{enumerate}

\section{Experimental Analysis}
\label{Sec:Experimental Analysis}
Here, we discuss the setup followed in experiments  and analyze the  performance of the proposed oblique and rotation double random models and baseline models or existing models (here, standard RaF \cite{breiman2001random}, standard DRaF \cite{han2020double}, MPRaF-T \cite{zhang2014oblique}, MPRaF-P \cite{zhang2014oblique}, MPRaF-N \cite{zhang2014oblique}, RaF-PCA \cite{zhang2014random} and RaF-LDA \cite{zhang2014random}). 

\subsection{Experimental Setup}
We evaluated the classification models  on UCI datasets \cite{Dua:2019} and real world fisheries datasets \cite{gonzalez2013exhaustive}. We follow the preprocessing scripts of  \cite{klambauer2017self} wherein the partitions of the training and testing sets are publicly available for evaluation. Table $1$ of the supplementary file summarizes the details of the $121$ datasets used for evaluation. The sample size of the datasets varies from $10$ to $130064$. Also, the dimensions of the feature samples vary from $3$ to $262$ and the number of classes vary from $2$ to $100$.

In all the ensemble models, $50$ is the number of base learners. At each non-terminal node, we  evaluated $\sqrt{n}$ number of features, here $n$ is the dimension of feature set and the minleaf parameter is set to default. We used CART \cite{breiman1984classification} as the base classifier.

\subsection{Statistical Analysis}
Table \ref{table:averageRaF} summarizes the classification performance of each ensemble model on $121$ datasets. From the given table, it is evident  that the average accuracy of the proposed DRaF-LDA, MPDRaF-P and DRaF-PCA are superior compared to the existing classifiers. Following \cite{fernandez2014we}, we rank each classifier based on its performance on each dataset. Every classifier in Friedman test is given a rank on a dataset with the worse performing classifier assigned higher rank and vice versa. Hence, a lower rank indicates better generalization performance of the model. The average rank of each classification model is presented in Table \ref{tab:average Friedman Rank}. 
It is evident that the average rank of the proposed ensemble models DRaF-LDA, DRaF-PCA, and MPDRaF-P is better as compared to all the existing classifiers. Furthermore, the rank of the proposed MPDRaF-T is better in comparison to existing classifiers (except standard DRaF and DRaF-LDA).

For evaluation of the models via statistical tests, we perform statistical analysis. We used Friedman test \cite{demvsar2006statistical} with corresponding Nemenyi post hoc test for the comparison of the models. Let $r_i^j$ be the the rank of the $j^{th}$ classification model assigned on the $i^{th}$ dataset among the $N$ datasets. In the Friedman test, average rank $\sum_i r^j_i$ is used for the evaluation of the classification models. When the number of datasets $(N)$ and the number of classifiers $(n)$ are large enough, then the Friedman statistic given as:
\begin{align}
    \chi^2_F=\frac{12N}{n(n+1)}\Bigg[ \sum_j R_j^2-\frac{n(n+1)^2}{4} \Bigg]
\end{align}
follows $\chi^2_F$ distribution with $(n-1)$ degrees of freedom under null hypothesis. As $\chi^2_F$ is undesirably conservative, hence, a better statistic is given as:
\begin{align}
    F_F=\frac{(N-1)\chi^2_F}{N(n-1)-\chi^2_F}
\end{align}
follows $F-$distribution with $(n-1)$ and $(n-1)(N-1)$ degrees of freedom. Under the null hypothesis, all the classifiers are  equal, hence, the  ranks of the classifiers are equal. If the null-hypothesis fails, Nemenyi post-hoc test \cite{nemenyi1962distribution} gives pairwise performance evaluation of the classifiers. Two classifiers are significantly different if their average ranks differ by at least the critical difference:
\begin{align}
    CD=q_{\alpha}\sqrt{\frac{n(n+1)}{6N}}
\end{align}
where $\alpha$ is the level of significance and $q_{\alpha}$ is the studentized range statistic divided by $\sqrt{2}$.

\begin{figure}
    \centering
    \includegraphics[width=0.8\textwidth]{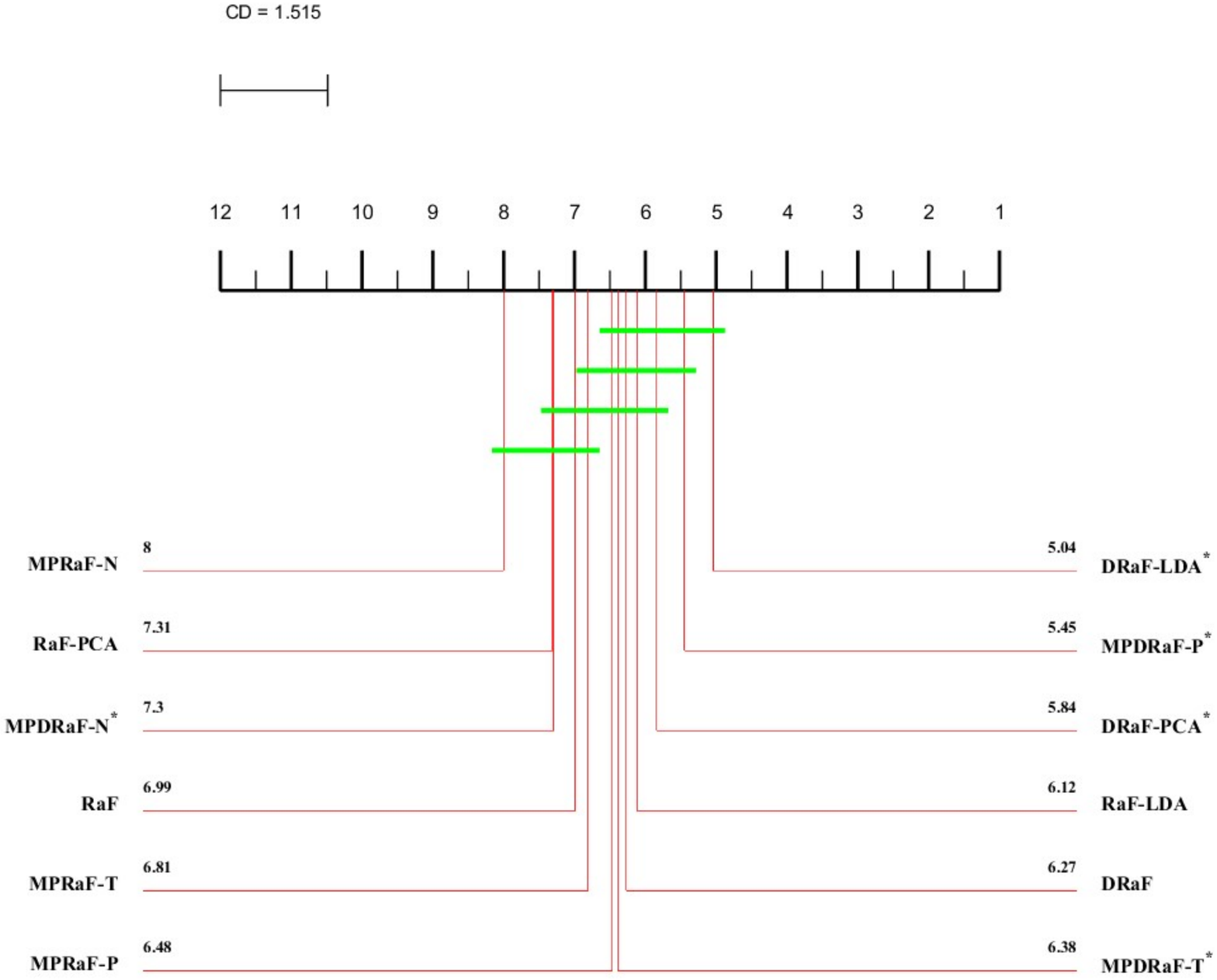}
    \caption{Nemenyi test based post hoc evaluation of classification models at $\alpha=5\%$ level of significance. The classification models which are not statistically different are connected.}
    \label{fig:Nemenyi figure}
\end{figure}

The average ranks of the classification models RaF,	MPRaF-T,	MPRaF-P,	MPRaF-N,	RaF-PCA,	RaF-LDA,	DRaF,	MPDRaF-T,	MPDRaF-P,	MPDRaF-N,	DRaF-PCA and	DRaF-LDA are $6.99,	6.81,	6.48,	8,	7.31,	6.12,	6.27,	6.38,	5.45,	7.3,	5.84$ and 	$5.04$ respectively. With simple calculations, we get $\chi^2_F=71.0559$ and $F_F=6.7675$. At $5\%$ level of significance i.e. $\alpha=5\%$, $F_F$ follows $F$-distribution with $(n-1)=11$ and $(n-1)(N-1)=1320$. From Statistical table, $F_F(11,1320)=1.8$. Since $6.7675>1.8$, hence we reject the null hypothesis. Thus, significant difference exists among the classification models. To get the significant difference, we use Nemenyi post hoc test. With simple calculations, critical difference $CD=1.5149$ with $q_{\alpha}=3.268$ at $5\%$ level of significance. From Figure \ref{fig:Nemenyi figure}, one can see the statistically significant difference exists among the models which are not connected by a line. Table \ref{tab:Nemenyi table} summarizes the Nemenyi post-hoc test results. From the table, it is evident that the proposed DRaF-LDA is significantly better in comparison to  RaF, MPRaF-T, MPRaF-N, RaF-PCA and MPDRaF-N classifiers. 
Also, the proposed DRaF-PCA is significantly better compared to the DRaF-PCA model.

The decision boundaries corresponding to the spiral dataset generated by the different classifiers are shown in Figure $5$ of the supplementary file.
\begin{table}[]
    \resizebox{\textwidth}{!}{
    \begin{tabular}{lcccccccccccc}
    \hline
       &RaF&MPRaF-T&MPRaF-P&MPRaF-N&RaF-PCA&RaF-LDA&DRaF&MPDRaF-T&MPDRaF-P&MPDRaF-N&DRaF-PCA&DRaF-LDA\\
       \hline
RaF&&&&&&&&&$r-$&&&$r-$\\
MPRaF-T&&&&&&&&&&&&$r-$\\
MPRaF-P&&&&$r+$&&&&&&&&\\
MPRaF-N&&&$r-$&&&$r-$&$r-$&$r-$&$r-$&&$r-$&$r-$\\
RaF-PCA&&&&&&&&&$r-$&&&$r-$\\
RaF-LDA&&&&$r+$&&&&&&&&\\
DRaF&&&&$r+$&&&&&&&&\\
MPDRaF-T&&&&$r+$&&&&&&&&\\
MPDRaF-P&$r+$&&&$r+$&$r+$&&&&&$r+$&&\\
MPDRaF-N&&&&&&&&&$r-$&&&$r-$\\
DRaF-PCA&&&&$r+$&&&&&&&&\\
DRaF-LDA&$r+$&$r+$&&$r+$&$r+$&&&&&$r+$&&\\
\hline
    \end{tabular}}
    Here, $r+$ denotes that the  the row model is significantly better than the column model. $r-$ denotes that the row model  is significantly worse than the  corresponding column model. Empty entries denote that no significant difference exists among the models of a cell.
    
    \caption{Significance difference of classification performance of the baseline models and the proposed oblique and rotation double random forest with Nemenyi posthoc tests based on the accuracy. }
    \label{tab:Nemenyi table}
\end{table}
\begin{landscape}
\begin{footnotesize}
\begin{longtable}[t]{|p{0.25\textwidth}|p{0.06\textwidth}|p{0.06\textwidth}|p{0.06\textwidth}|p{0.06\textwidth}|p{0.1\textwidth}|p{0.1\textwidth}|c|c|c|c|c|c|p{0.1\textwidth}|}
\caption[this is]{Classification accuracy 
of RaF \cite{breiman2001random}, MPRaF-T \cite{zhang2014oblique}, MPRaF-P \cite{zhang2014oblique}, MPRaF-N \cite{zhang2014oblique}, RaF-PCA \cite{zhang2014random}, RaF-LDA \cite{zhang2014random}, DRaF \cite{han2020double}, MPDRaF-T, MPDRaF-P, MPDRaF-N, DRaF-PCA AND DRaF-LDA classification models } \label{table:averageRaF} \\
\hline
Datasets &RaF &MPRaF-T&MPRaF-P&MPRaF-N&RaF-PCA&RaF-LDA&DRaF&MPDRaF-T$^*$&MPDRaF-P$^*$&MPDRaF-N$^*$&DRaF-PCA$^*$&DRaF-LDA$^*$\\
\hline 
\endfirsthead
\multicolumn{13}{c}%
{{\bfseries \tablename\ \thetable{} -- continued from previous page}} \\
\hline
Datasets &RaF &MPRaF-T&MPRaF-P&MPRaF-N&RaF-PCA&RaF-LDA&DRaF&MPDRaF-T$^*$&MPDRaF-P$^*$&MPDRaF-N$^*$&DRaF-PCA$^*$&DRaF-LDA$^*$\\
 \hline 
\endhead
\hline \multicolumn{13}{|r|}{{Continued on next page}} \\ \hline
\endfoot
\endlastfoot

\hline
   abalone&$64.68$&$64.99$&$65.54$&$65.06$&$64.85$&$65.4$&$64.18$&$65.33$&$63.94$&$65.33$&$64.06$&$65.35$\\
acute-inflammation&$100$&$100$&$100$&$100$&$100$&$100$&$100$&$100$&$100$&$100$&$100$&$100$\\
acute-nephritis&$100$&$100$&$100$&$100$&$100$&$100$&$100$&$100$&$100$&$100$&$100$&$100$\\
adult&$85.79$&$85.04$&$85.54$&$84.5$&$85.6$&$85.47$&$85.63$&$85.12$&$85.45$&$84.41$&$85.58$&$85.47$\\
annealing&$54.25$&$76$&$38.75$&$76$&$62.25$&$65$&$37$&$76$&$56$&$76$&$65.25$&$64$\\
arrhythmia&$73.01$&$63.05$&$73.23$&$61.5$&$65.49$&$67.04$&$73.67$&$63.05$&$73.89$&$60.62$&$70.35$&$70.35$\\
audiology-std&$75$&$70$&$76$&$24$&$55$&$48$&$78$&$59$&$78$&$25$&$60$&$58$\\
balance-scale&$86.7$&$89.42$&$88.94$&$89.42$&$88.62$&$89.42$&$82.69$&$87.82$&$86.86$&$89.58$&$85.42$&$86.38$\\
balloons&$81.25$&$87.5$&$87.5$&$93.75$&$81.25$&$75$&$87.5$&$93.75$&$81.25$&$81.25$&$81.25$&$87.5$\\
bank&$89.6$&$88.61$&$89.2$&$88.63$&$89.45$&$89.91$&$89.93$&$88.87$&$89.29$&$88.83$&$89.6$&$89.82$\\
blood&$76.6$&$76.74$&$77.27$&$77.81$&$76.6$&$77.01$&$75.67$&$77.14$&$75.94$&$77.14$&$75.8$&$76.34$\\
breast-cancer&$73.94$&$73.94$&$73.94$&$73.94$&$76.06$&$76.76$&$75.35$&$73.24$&$74.65$&$76.06$&$72.89$&$75$\\
breast-cancer-wisc&$97.29$&$97.71$&$97.43$&$97$&$97.14$&$97.43$&$97.14$&$97.86$&$97.71$&$97.57$&$97.43$&$97.29$\\
breast-cancer-wisc-diag&$95.6$&$96.83$&$96.83$&$97.71$&$95.6$&$97.01$&$95.77$&$97.01$&$96.65$&$96.3$&$96.83$&$97.01$\\
breast-cancer-wisc-prog&$80.1$&$80.61$&$79.59$&$81.12$&$80.1$&$80.61$&$81.63$&$82.65$&$82.14$&$83.67$&$82.14$&$82.14$\\
breast-tissue&$70.19$&$69.23$&$71.15$&$71.15$&$73.08$&$75$&$73.08$&$69.23$&$69.23$&$68.27$&$73.08$&$70.19$\\
car&$96.93$&$95.31$&$96.99$&$88.08$&$96.76$&$96.76$&$97.05$&$95.37$&$97.8$&$87.96$&$97.16$&$98.15$\\
cardiotocography-10clases&$86.11$&$82.44$&$85.59$&$79.8$&$84.37$&$84.84$&$87.15$&$83.47$&$86.53$&$81.87$&$84.93$&$85.64$\\
cardiotocography-3clases&$94.02$&$92.75$&$94.26$&$91.57$&$92.33$&$93.27$&$94.92$&$93.22$&$94.73$&$92.23$&$92.7$&$93.69$\\
chess-krvk&$69.6$&$65.97$&$70.12$&$52.02$&$73.33$&$71.72$&$69.92$&$66.34$&$71.36$&$52.22$&$75.18$&$73.24$\\
chess-krvkp&$98.25$&$97.97$&$98.62$&$97.43$&$98.25$&$98.59$&$98.56$&$98.53$&$98.94$&$97.84$&$98.75$&$98.81$\\
congressional-voting&$62.39$&$61.24$&$61.01$&$61.24$&$60.55$&$61.01$&$61.7$&$62.39$&$61.7$&$60.78$&$60.78$&$61.7$\\
conn-bench-sonar-mines-rocks&$76.92$&$78.37$&$78.85$&$78.37$&$76.44$&$78.85$&$79.33$&$77.4$&$80.77$&$79.33$&$80.77$&$84.13$\\
conn-bench-vowel-deterding&$98.48$&$99.78$&$99.13$&$99.62$&$99.57$&$99.46$&$98.97$&$100$&$99.73$&$99.68$&$99.95$&$99.95$\\
connect-4&$83.54$&$76$&$81.2$&$75.41$&$82.63$&$82.31$&$84.01$&$75.91$&$81.7$&$75.4$&$83.37$&$82.83$\\
contrac&$53.94$&$50.41$&$53.13$&$49.8$&$51.9$&$50.68$&$53.33$&$48.78$&$51.15$&$52.31$&$51.15$&$51.56$\\
credit-approval&$87.5$&$86.92$&$86.05$&$88.08$&$88.08$&$87.5$&$87.5$&$86.63$&$87.21$&$85.61$&$87.21$&$87.06$\\
cylinder-bands&$81.25$&$76.17$&$80.47$&$73.05$&$78.71$&$77.73$&$82.03$&$76.95$&$82.03$&$79.3$&$80.86$&$80.47$\\
dermatology&$98.35$&$98.08$&$98.35$&$96.7$&$97.8$&$97.8$&$98.08$&$97.8$&$97.8$&$97.8$&$97.53$&$97.25$\\
echocardiogram&$84.85$&$84.85$&$85.61$&$84.09$&$84.09$&$83.33$&$84.09$&$85.61$&$84.85$&$84.85$&$84.09$&$84.09$\\
ecoli&$86.31$&$87.2$&$88.99$&$87.8$&$87.5$&$87.5$&$88.1$&$86.01$&$88.1$&$88.39$&$86.61$&$86.61$\\
energy-y1&$94.79$&$92.58$&$94.92$&$94.14$&$94.53$&$95.7$&$95.83$&$92.19$&$95.96$&$94.27$&$96.09$&$96.22$\\
energy-y2&$89.06$&$89.45$&$89.84$&$89.32$&$89.97$&$89.71$&$88.28$&$89.06$&$88.8$&$89.71$&$89.84$&$89.45$\\
fertility&$88$&$89$&$89$&$88$&$88$&$88$&$88$&$88$&$88$&$88$&$88$&$88$\\
flags&$67.19$&$55.21$&$64.58$&$56.77$&$56.25$&$57.29$&$66.67$&$54.69$&$63.02$&$54.69$&$62.5$&$62.5$\\
glass&$73.11$&$69.34$&$75.47$&$70.28$&$70.75$&$72.17$&$76.89$&$69.34$&$75.47$&$67.92$&$71.7$&$74.06$\\
haberman-survival&$71.05$&$71.71$&$71.05$&$72.37$&$70.07$&$71.38$&$69.74$&$70.39$&$69.08$&$70.72$&$69.41$&$68.75$\\
hayes-roth&$87.5$&$86.61$&$84.82$&$81.25$&$89.29$&$87.5$&$89.29$&$85.71$&$90.18$&$75$&$89.29$&$88.39$\\
heart-cleveland&$57.89$&$61.51$&$57.57$&$59.21$&$58.22$&$59.21$&$55.59$&$59.21$&$58.22$&$59.54$&$60.2$&$57.57$\\
heart-hungarian&$83.9$&$84.93$&$84.25$&$84.25$&$84.59$&$84.59$&$84.25$&$83.56$&$83.56$&$85.27$&$84.59$&$84.59$\\
heart-switzerland&$41.13$&$43.55$&$41.13$&$44.35$&$43.55$&$45.16$&$41.94$&$39.52$&$41.94$&$41.13$&$45.97$&$47.58$\\
heart-va&$35.5$&$34.5$&$35.5$&$36.5$&$34$&$37.5$&$36$&$32.5$&$36.5$&$39.5$&$33.5$&$34.5$\\
hepatitis&$83.33$&$82.05$&$82.05$&$86.54$&$82.69$&$84.62$&$82.69$&$82.69$&$81.41$&$82.69$&$81.41$&$80.77$\\
hill-valley&$53.84$&$66.75$&$63$&$65.88$&$64.03$&$66.25$&$54.17$&$70.09$&$66.79$&$66.58$&$67.2$&$66.75$\\
horse-colic&$86.4$&$86.03$&$87.87$&$87.5$&$82.35$&$85.29$&$86.76$&$83.46$&$86.76$&$84.56$&$80.51$&$81.62$\\
ilpd-indian-liver&$71.4$&$70.72$&$71.23$&$71.23$&$73.29$&$71.23$&$71.23$&$72.6$&$72.95$&$71.23$&$73.46$&$71.75$\\
image-segmentation&$93.8$&$94.18$&$94.75$&$92.46$&$94.96$&$95.07$&$94.85$&$94.63$&$95.15$&$92.57$&$95.93$&$96.06$\\
ionosphere&$91.76$&$93.75$&$93.47$&$93.18$&$94.03$&$94.6$&$91.48$&$93.75$&$94.03$&$94.32$&$94.89$&$93.18$\\
iris&$95.27$&$97.3$&$97.3$&$97.97$&$95.95$&$96.62$&$95.95$&$97.3$&$95.95$&$97.3$&$96.62$&$96.62$\\
led-display&$74.3$&$72$&$73.7$&$72.4$&$73.9$&$73.6$&$71.7$&$72.1$&$72.4$&$71.2$&$71.6$&$72.1$\\
lenses&$83.33$&$79.17$&$83.33$&$79.17$&$79.17$&$87.5$&$83.33$&$79.17$&$75$&$83.33$&$79.17$&$79.17$\\
letter&$95.31$&$95.35$&$95.18$&$94.83$&$94.74$&$95.62$&$95.86$&$95.85$&$96$&$95.02$&$96.06$&$96.73$\\
libras&$76.94$&$84.17$&$79.17$&$79.44$&$80.28$&$81.39$&$79.72$&$86.11$&$84.72$&$86.67$&$85.28$&$85.83$\\
low-res-spect&$90.79$&$91.17$&$91.35$&$89.47$&$90.6$&$91.54$&$91.54$&$91.17$&$91.73$&$90.79$&$91.17$&$91.35$\\
lung-cancer&$46.88$&$46.88$&$50$&$53.13$&$40.63$&$43.75$&$50$&$31.25$&$53.13$&$50$&$46.88$&$50$\\
lymphography&$79.05$&$85.14$&$83.11$&$83.78$&$84.46$&$84.46$&$85.81$&$86.49$&$83.11$&$85.81$&$84.46$&$86.49$\\
magic&$87.01$&$86.37$&$86.26$&$86.69$&$87.38$&$87.06$&$87.19$&$86.51$&$86.78$&$86.88$&$87.79$&$87.57$\\
mammographic&$81.98$&$81.67$&$80.63$&$81.67$&$80.63$&$80.42$&$79.9$&$81.46$&$80.63$&$81.35$&$80.1$&$80.42$\\
miniboone&$93.33$&$93.07$&$93.24$&$92.76$&$93.21$&$93.46$&$93.69$&$93.5$&$93.64$&$93.3$&$93.65$&$93.88$\\
molec-biol-promoter&$84.62$&$79.81$&$84.62$&$82.69$&$71.15$&$78.85$&$91.35$&$84.62$&$87.5$&$80.77$&$83.65$&$85.58$\\
molec-biol-splice&$94.2$&$86.57$&$93.1$&$85.01$&$84.1$&$89.9$&$94.7$&$87.23$&$93.22$&$87.14$&$87.05$&$90.56$\\
monks-1&$59.95$&$60.59$&$58.39$&$57.52$&$58.04$&$58.16$&$60.65$&$60.47$&$58.8$&$58.97$&$58.22$&$59.32$\\
monks-2&$66.78$&$66.9$&$66.9$&$67.01$&$66.84$&$67.01$&$66.55$&$66.96$&$66.61$&$67.13$&$66.9$&$66.72$\\
monks-3&$53.01$&$56.6$&$52.78$&$54.34$&$53.36$&$52.89$&$52.78$&$54.17$&$52.95$&$52.78$&$53.01$&$53.76$\\
mushroom&$100$&$100$&$100$&$100$&$100$&$100$&$100$&$100$&$100$&$100$&$100$&$100$\\
musk-1&$86.13$&$86.97$&$83.82$&$87.18$&$86.13$&$83.82$&$86.34$&$89.29$&$86.97$&$87.82$&$88.24$&$85.92$\\
musk-2&$97.21$&$96.12$&$95.94$&$95.69$&$95.98$&$96.12$&$98.12$&$96.53$&$96.45$&$96.07$&$96.71$&$96.95$\\
nursery&$99.28$&$98.58$&$99.21$&$96.74$&$99.22$&$99.33$&$99.31$&$98.9$&$99.53$&$96.95$&$99.66$&$99.76$\\
OM\_nucleus\_4d&$77.65$&$81.57$&$82.55$&$80.69$&$82.55$&$82.65$&$79.8$&$84.31$&$83.24$&$82.84$&$82.45$&$84.31$\\
OM\_states\_2f&$91.37$&$91.57$&$92.16$&$91.86$&$91.86$&$92.16$&$92.35$&$92.35$&$92.35$&$92.55$&$92.06$&$92.45$\\
OT\_nucleus\_2f&$79.39$&$81.58$&$82.79$&$83.11$&$82.57$&$82.24$&$80.7$&$83.44$&$82.79$&$82.89$&$83.77$&$83.22$\\
OT\_states\_5b&$90.9$&$92.54$&$92$&$92.65$&$92.65$&$93.31$&$92.21$&$93.64$&$93.09$&$93.09$&$92.87$&$93.64$\\
optical&$96.08$&$95.66$&$96.26$&$84.65$&$95.72$&$91.62$&$96.91$&$96.37$&$96.74$&$83.85$&$96.59$&$94.3$\\
ozone&$97.08$&$97.2$&$97.16$&$97.16$&$97.16$&$97.16$&$97.08$&$97.16$&$97.16$&$97.16$&$97.16$&$97.2$\\
page-blocks&$97.08$&$96.98$&$97.3$&$96.78$&$97.09$&$97.13$&$97.08$&$97.08$&$97.09$&$97.08$&$97.09$&$97.28$\\
parkinsons&$88.78$&$92.35$&$89.8$&$91.84$&$87.76$&$90.82$&$90.31$&$92.86$&$92.35$&$92.35$&$90.82$&$91.84$\\
pendigits&$95.05$&$96.76$&$95.75$&$96.06$&$96.38$&$96.48$&$95.48$&$96.96$&$96.22$&$96.18$&$96.53$&$96.58$\\
pima&$76.69$&$75.26$&$75.52$&$75.13$&$74.61$&$74.87$&$73.96$&$74.74$&$74.22$&$74.48$&$74.09$&$75$\\
pittsburg-bridges-MATERIAL&$91.35$&$93.27$&$91.35$&$92.31$&$92.31$&$92.31$&$88.46$&$93.27$&$89.42$&$92.31$&$90.38$&$91.35$\\
pittsburg-bridges-REL-L&$74.04$&$75.96$&$73.08$&$75.96$&$75$&$73.08$&$73.08$&$74.04$&$71.15$&$78.85$&$73.08$&$75$\\
pittsburg-bridges-SPAN&$61.96$&$72.83$&$63.04$&$67.39$&$69.57$&$71.74$&$60.87$&$67.39$&$61.96$&$67.39$&$66.3$&$66.3$\\
pittsburg-bridges-T-OR-D&$88$&$88$&$88$&$88$&$88$&$90$&$89$&$88$&$88$&$88$&$88$&$88$\\
pittsburg-bridges-TYPE&$68.27$&$69.23$&$67.31$&$66.35$&$71.15$&$66.35$&$67.31$&$69.23$&$68.27$&$69.23$&$70.19$&$71.15$\\
planning&$70$&$67.78$&$70$&$70$&$70.56$&$69.44$&$69.44$&$71.67$&$71.11$&$72.22$&$70.56$&$70.56$\\
plant-margin&$79.25$&$75.06$&$72.25$&$72.56$&$76$&$75.5$&$81.56$&$76.88$&$77.5$&$73.06$&$79.75$&$82.44$\\
plant-shape&$59.44$&$66.13$&$62.13$&$65.25$&$65.31$&$68.44$&$61.94$&$67.75$&$66.31$&$66.13$&$70$&$73.44$\\
plant-texture&$77.94$&$77.25$&$76.06$&$75.06$&$75.81$&$76.81$&$80.56$&$79.13$&$79.06$&$75.56$&$79$&$81.63$\\
post-operative&$72.73$&$71.59$&$70.45$&$71.59$&$69.32$&$67.05$&$70.45$&$72.73$&$69.32$&$68.18$&$69.32$&$68.18$\\
primary-tumor&$54.88$&$51.83$&$55.18$&$52.44$&$53.05$&$56.1$&$54.88$&$53.05$&$53.66$&$53.35$&$54.57$&$54.57$\\
ringnorm&$95.19$&$90.41$&$90.81$&$90.85$&$97.01$&$97.09$&$95.46$&$91.99$&$92.15$&$92.54$&$97.24$&$97.15$\\
seeds&$93.27$&$94.71$&$91.83$&$91.83$&$93.75$&$92.31$&$93.75$&$93.75$&$95.19$&$92.31$&$92.79$&$93.75$\\
semeion&$92.4$&$89.51$&$91.52$&$89.13$&$88.69$&$91.96$&$92.46$&$89.7$&$92.09$&$90.52$&$91.14$&$93.22$\\
soybean&$90.29$&$89.23$&$90.56$&$82.71$&$89.83$&$86.3$&$90.36$&$88.63$&$90.76$&$83.38$&$90.36$&$87.43$\\
spambase&$94.39$&$94.5$&$94.15$&$94.11$&$94.8$&$94.48$&$94.72$&$94.91$&$94.83$&$94.3$&$95.3$&$95.17$\\
spect&$68.95$&$61.56$&$65.46$&$59.68$&$60.75$&$61.29$&$65.05$&$60.75$&$63.04$&$60.22$&$61.42$&$62.1$\\
spectf&$91.98$&$91.98$&$91.98$&$91.98$&$91.98$&$91.84$&$91.98$&$91.84$&$91.98$&$91.98$&$91.98$&$91.84$\\
statlog-australian-credit&$67.3$&$65.26$&$66.57$&$67.15$&$63.66$&$63.23$&$64.39$&$63.37$&$65.55$&$63.52$&$64.53$&$63.08$\\
statlog-german-credit&$77.5$&$74.8$&$75.4$&$73.9$&$75.3$&$77.7$&$77.4$&$73.9$&$75.9$&$72.8$&$76.1$&$76.1$\\
statlog-heart&$85.45$&$87.31$&$86.19$&$86.19$&$85.45$&$85.45$&$85.07$&$85.82$&$85.45$&$83.21$&$85.45$&$85.45$\\
statlog-image&$97.27$&$97.66$&$97.57$&$96.66$&$97.88$&$97.92$&$97.88$&$97.92$&$98.31$&$97.18$&$98.09$&$98.27$\\
statlog-landsat&$89.94$&$89.99$&$89.99$&$89.04$&$89.78$&$89.88$&$90.78$&$90.89$&$90.73$&$89.44$&$90.76$&$90.98$\\
statlog-shuttle&$99.96$&$99.87$&$99.95$&$99.76$&$99.94$&$99.96$&$99.99$&$99.9$&$99.97$&$99.78$&$99.97$&$99.97$\\
statlog-vehicle&$73.58$&$76.3$&$77.73$&$75.95$&$78.08$&$79.03$&$75.71$&$76.18$&$77.61$&$77.25$&$78.32$&$80.45$\\
steel-plates&$78.04$&$78.04$&$76.75$&$75.15$&$75.05$&$76.49$&$78.4$&$78.2$&$78.76$&$76.86$&$77.94$&$77.99$\\
synthetic-control&$97.67$&$99.83$&$98.5$&$98.33$&$97.17$&$99.17$&$98.5$&$99.33$&$99.33$&$98.83$&$98.5$&$99.67$\\
teaching&$59.21$&$58.55$&$60.53$&$57.24$&$55.92$&$60.53$&$58.55$&$58.55$&$59.87$&$57.24$&$59.21$&$59.87$\\
thyroid&$98.88$&$95.86$&$98.89$&$93.26$&$98.7$&$97.65$&$98.96$&$96.05$&$98.93$&$93.46$&$98.87$&$98.16$\\
tic-tac-toe&$97.91$&$97.49$&$97.7$&$94.77$&$97.07$&$98.01$&$98.64$&$98.95$&$98.85$&$98.22$&$98.43$&$99.06$\\
titanic&$78.95$&$78.68$&$78.95$&$78.32$&$78.95$&$78.95$&$78.95$&$78.95$&$78.95$&$78.5$&$78.95$&$78.95$\\
trains&$87.5$&$100$&$87.5$&$87.5$&$87.5$&$87.5$&$87.5$&$87.5$&$87.5$&$87.5$&$87.5$&$87.5$\\
twonorm&$96.8$&$97.59$&$97.68$&$97.57$&$97.68$&$97.55$&$96.8$&$97.57$&$97.53$&$97.42$&$97.68$&$97.66$\\
vertebral-column-2clases&$83.77$&$86.69$&$86.04$&$86.04$&$85.06$&$86.69$&$82.14$&$86.36$&$86.36$&$87.01$&$83.44$&$85.06$\\
vertebral-column-3clases&$83.44$&$84.09$&$83.44$&$83.77$&$83.77$&$83.77$&$84.74$&$86.04$&$84.42$&$85.71$&$85.39$&$86.36$\\
wall-following&$99.3$&$94.24$&$98.41$&$93.71$&$96.17$&$96.19$&$99.52$&$94.54$&$98.57$&$94.68$&$96.87$&$96.92$\\
waveform&$84.54$&$85.4$&$85.04$&$85.44$&$84.8$&$85.4$&$83.76$&$85.26$&$85.9$&$85.48$&$85.12$&$85.78$\\
waveform-noise&$85.5$&$85.2$&$86.24$&$85.74$&$85.08$&$85.84$&$85.22$&$85.44$&$85.44$&$85.52$&$85.6$&$86.14$\\
wine&$97.73$&$98.86$&$99.43$&$97.73$&$97.16$&$98.86$&$97.73$&$97.73$&$99.43$&$98.3$&$97.16$&$99.43$\\
wine-quality-red&$65.81$&$68$&$67.38$&$68.19$&$68.31$&$67.56$&$68$&$67.5$&$69.19$&$67.31$&$68.81$&$67.88$\\
wine-quality-white&$67.01$&$67.28$&$67.57$&$66.14$&$67.87$&$67.69$&$68.2$&$67.85$&$67.97$&$66.91$&$68.57$&$68.4$\\
yeast&$61.52$&$62.06$&$61.79$&$62.2$&$62.53$&$62.53$&$60.58$&$61.39$&$61.39$&$60.98$&$60.98$&$61.79$\\
zoo&$99$&$99$&$98$&$98$&$99$&$99$&$98$&$97$&$98$&$99$&$98$&$98$\\
\hline
Average Accuracy&$81.86$&$81.98$&$81.96$&$80.83$&$81.48$&$81.9$&$82.09$&$81.79$&$82.39$&$80.96$&$82.2$&$82.55$\\
\hline
\multicolumn{13}{l}{Here, $^*$ denotes the methods introduced in this paper.}\\
\multicolumn{13}{l}{OM denotes oocytes\_merluccius, OT denotes oocytes\_trisopterus.}
    \end{longtable}
    \end{footnotesize}
\end{landscape}

\begin{table}[h]
    \centering
    \begin{tabular}{lcccc}
    \hline
        &Rank&Average Rank&Average Accuracy& Average Time(s)\\
        \hline
DRaF-LDA$^*$&$1$&$5.04$&$82.55$&$758.68$\\
MPDRaF-P$^*$&$2$&$5.45$&$82.39$&$80.83$\\
DRaF-PCA$^*$&$3$&$5.84$&$82.2$&$765.81$\\
RaF-LDA&$4$&$6.12$&$81.9$&$732.63$\\
DRaF&$5$&$6.27$&$82.09$&$523.32$\\
MPDRaF-T$^*$&$6$&$6.38$&$81.79$&$30.66$\\
MPRaF-P&$7$&$6.48$&$81.96$&$56.3$\\
MPRaF-T&$8$&$6.81$&$81.98$&$24.64$\\
RaF&$9$&$6.99$&$81.86$&$383.43$\\
MPDRaF-N$^*$&$10$&$7.3$&$80.96$&$32.12$\\
RaF-PCA&$11$&$7.31$&$81.48$&$719.97$\\
MPRaF-N&$12$&$8$&$80.83$&$26.76$\\
\hline
\multicolumn{4}{l}{Here $^*$ denotes the methods introduced in this paper.}
    \end{tabular}
    
    \caption{Overall comparison of the baseline classification models, proposed oblique and rotation double random forest models.}
    \label{tab:average Friedman Rank}
\end{table}

\begin{landscape}

\begin{table}[]
    \resizebox{\linewidth}{!}{
    \begin{tabular}{lcccccccccccc}
    \hline
    &RaF&MPRaF-T&MPRaF-P&MPRaF-N&RaF-PCA&RaF-LDA&DRaF&MPDRaF-T$^*$&MPDRaF-P$^*$&MPDRaF-N$^*$&DRaF-PCA$^*$\\
MPRaF-T&$[57,10,54]$&&&&&&&&&&\\
MPRaF-P&$[60,17,44]$&$[60,13,48]$&&&&&&&&&\\
MPRaF-N&$[50,10,61]$&$[36,14,71]$&$[40,16,65]$&&&&&&&&\\
RaF-PCA&$[49,16,56]$&$[51,8,62]$&$[47,11,63]$&$[63,15,43]$&&&&&&&\\
RaF-LDA&$[67,10,44]$&$[63,10,48]$&$[61,14,46]$&$[77,13,31]$&$[70,17,34]$&&&&&&\\
DRaF&$[69,14,38]$&$[65,8,48]$&$[58,17,46]$&$[64,11,46]$&$[60,15,46]$&$[54,11,56]$&&&&&\\
MPDRaF-T$^*$&$[56,14,51]$&$[61,15,45]$&$[54,10,57]$&$[70,15,36]$&$[60,14,47]$&$[48,12,61]$&$[52,12,57]$&&&&\\
MPDRaF-P$^*$&$[69,13,39]$&$[74,10,37]$&$[70,14,37]$&$[76,12,33]$&$[68,18,35]$&$[67,9,45]$&$[63,16,42]$&$[59,16,46]$&&&\\
MPDRaF-N$^*$&$[46,14,61]$&$[44,12,65]$&$[45,13,63]$&$[74,13,34]$&$[55,14,52]$&$[40,12,69]$&$[47,14,60]$&$[36,17,68]$&$[37,13,71]$&&\\
DRaF-PCA$^*$&$[61,14,46]$&$[65,10,46]$&$[56,16,49]$&$[77,11,33]$&$[72,22,27]$&$[64,14,43]$&$[57,14,50]$&$[59,10,52]$&$[49,20,52]$&$[70,11,40]$&\\
DRaF-LDA$^*$&$[69,11,41]$&$[70,10,41]$&$[62,15,44]$&$[79,11,31]$&$[71,16,34]$&$[77,12,32]$&$[67,12,42]$&$[68,15,38]$&$[62,13,46]$&$[79,8,34]$&$[66,22,33]$\\
\hline
    \end{tabular}}
         Here, $^*$ denotes the proposed methods, $[a,b,c]$ entry in each cell  denotes that row method wins $a$-times, loses $c$-times and ties $b$-times with respect to  column method.
    \caption{Pairwise win-tie-loss count}
    \label{tab:pairwise win-tie-loss}
\end{table}
\end{landscape}

\begin{landscape}
\begin{table}[]
    \resizebox{\linewidth}{!}{
    \begin{tabular}{lccccccccccccc}
    \hline
      &RaF&MPRaF-T&MPRaF-P&MPRaF-N&RaF-PCA&RaF-LDA&DRaF&MPDRaF-T$^*$&MPDRaF-P$^*$&MPDRaF-N$^*$&DRaF-PCA$^*$&DRaF-LDA$^*$\\
RaF&&&&&&$r-$&$r-$&&$r-$&&&$r-$\\
MPRaF-T&&&&&&&&&$r-$&&&$r-$\\
MPRaF-P&&&&&&&&&$r-$&&&\\
MPRaF-N&&&&&&$r-$&&$r-$&$r-$&$r-$&$r-$&$r-$\\
RaF-PCA&&&&&&$r-$&&&$r-$&&$r-$&$r-$\\
RaF-LDA&$r+$&&&$r+$&$r+$&&&&$r-$&&&$r-$\\
DRaF&$r+$&&&&&&&&&&&$r-$\\
MPDRaF-T$^*$&&&&$r+$&&&&&&&&$r-$\\
MPDRaF-P$^*$&$r+$&$r+$&$r+$&$r+$&$r+$&$r+$&&&&&&\\
MPDRaF-N$^*$&&&&$r+$&&&&&&&$r-$&$r-$\\
DRaF-PCA$^*$&&&&$r+$&$r+$&&&&&$r+$&&$r-$\\
DRaF-LDA$^*$&$r+$&$r+$&&$r+$&$r+$&$r+$&$r+$&$r+$&&$r+$&$r+$&\\
\hline
    \end{tabular}}
      Here, $^*$ denotes the methods introduced in this paper, $r+$ denotes that the method in the corresponding row is significantly better as compared to the method given in the corresponding column. $r-$ denotes that the row method is significantly worse than the method given in the corresponding column. Blank entries denote that no significant difference exists among the methods in the cell's corresponding row and column.  
    \caption{Pairwise win-tie-loss: Sign test}
    \label{tab:pairwise win-tie-loss:Sign test}
\end{table}
\end{landscape}

 \subsection{Win-Tie-Loss: Sign test}
Under the null hypothesis, the pair of classifiers is significantly different if each  classification model wins $N/2$ in $N$ datasets. The number of  wins follow binomial distribution. When $N$ is large enough, the number of wins follow $N(N/2,\sqrt{N}/2),$ and hence, $z$-test can be used: two models are significantly better with $p<0.05$ if any model has least $N/2+1.96\sqrt{N}/2$  wins. Since tied matches favor of null hypothesis, hence, we split the number of ties between the models evenly and if the number is odd we ignore one.

Table \ref{tab:pairwise win-tie-loss} summarizes the count of win tie loss results among the given classification models. 
One can see that the proposed rotation double random forest (DRaF-PCA and DRaF-LDA) achieved more  wins as compared to the existing models. Compared to the existing MPRaF-N and RaF-PCA models, the proposed MPDRaF-N emerged as winner in more datasets. Also, the proposed MPDRaF-P model emerged as the winner in more datasets in comparison to the given baseline models. Table \ref{tab:pairwise win-tie-loss:Sign test} shows that the proposed DRaF-LDA model is significantly better as compared to the RaF, MPRaF-T, MPRaF-N, RaF-PCA, RaF-LDA and DRaF models. The proposed DRaF-PCA model is significantly better compared to the existing MPRaF-N and RaF-PCA models. Also, the proposed MPDRaF-P is significantly better as compared to the existing models except DRaF model. 

\subsection{Effect of ``mtry" parameter}
\begin{figure} 
\caption{Effect of the \textit{``mtry"} parameter.}
\label{fig:effect of mtry}
    \subcaptionbox{Echocardiogram 
    \label{fig:echocardiogram}} { %
      \includegraphics[width=\textwidth]{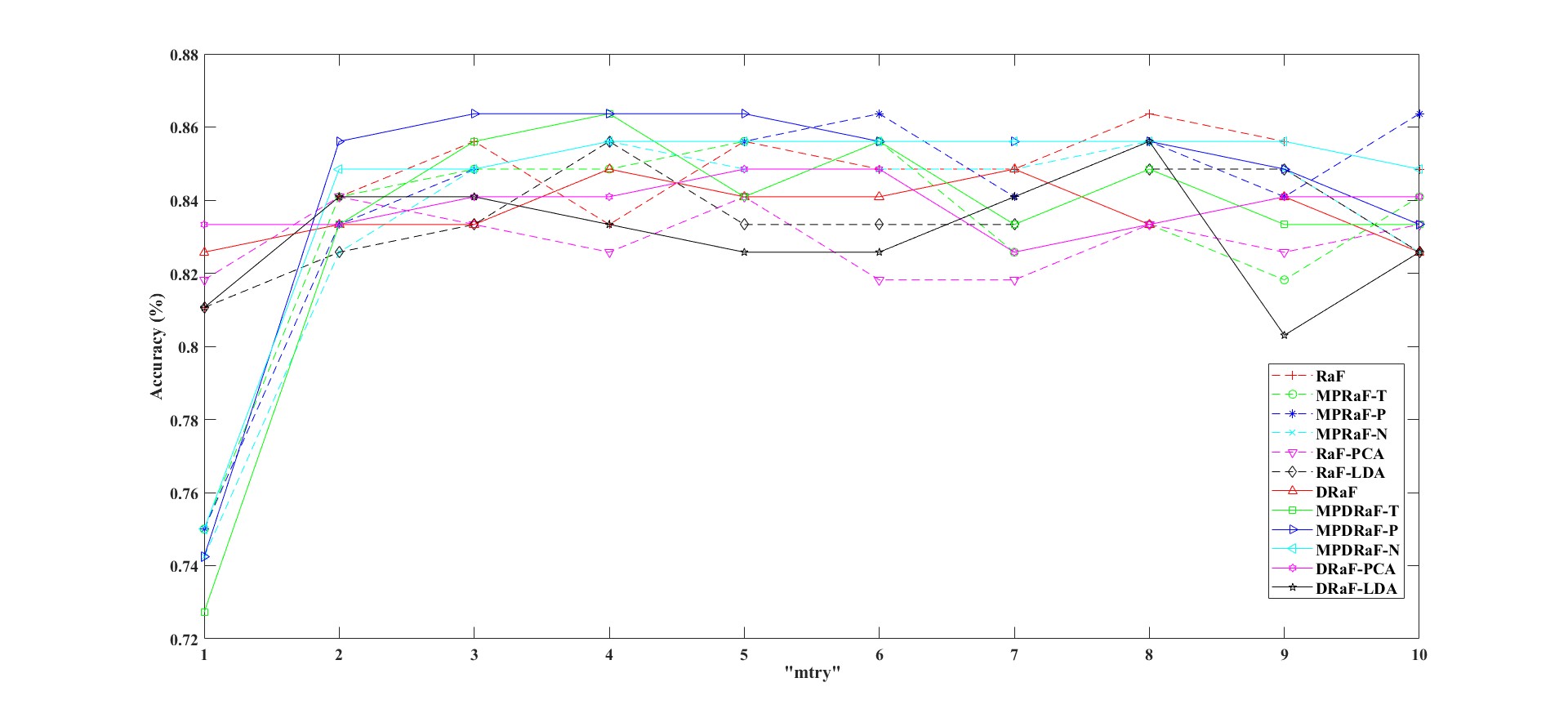}}\\
    \subcaptionbox{Ecoli \label{fig:ecoli}} { %
      \includegraphics[width=\textwidth]{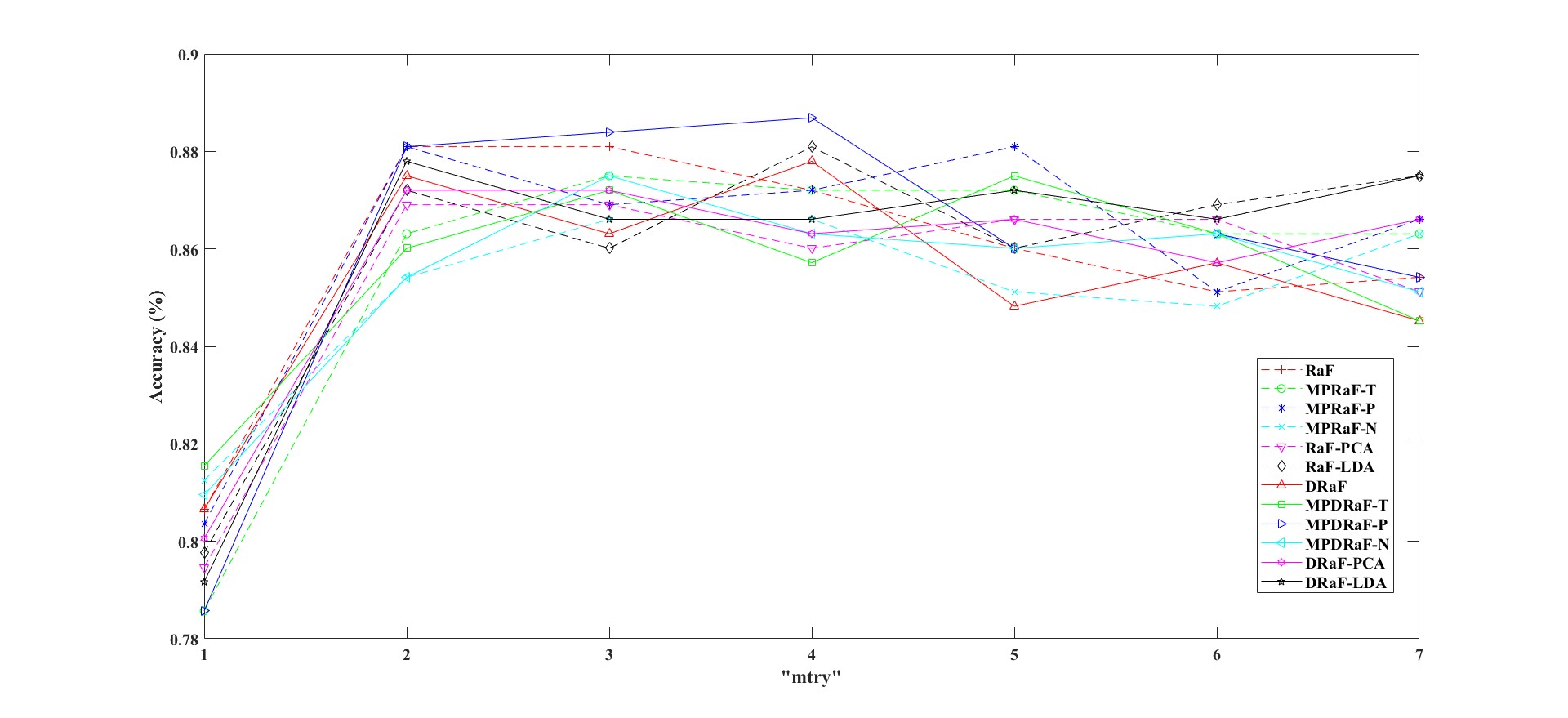}} \\
    \subcaptionbox{Soybean \label{fig:soybean}} { %
      \includegraphics[width=\textwidth]{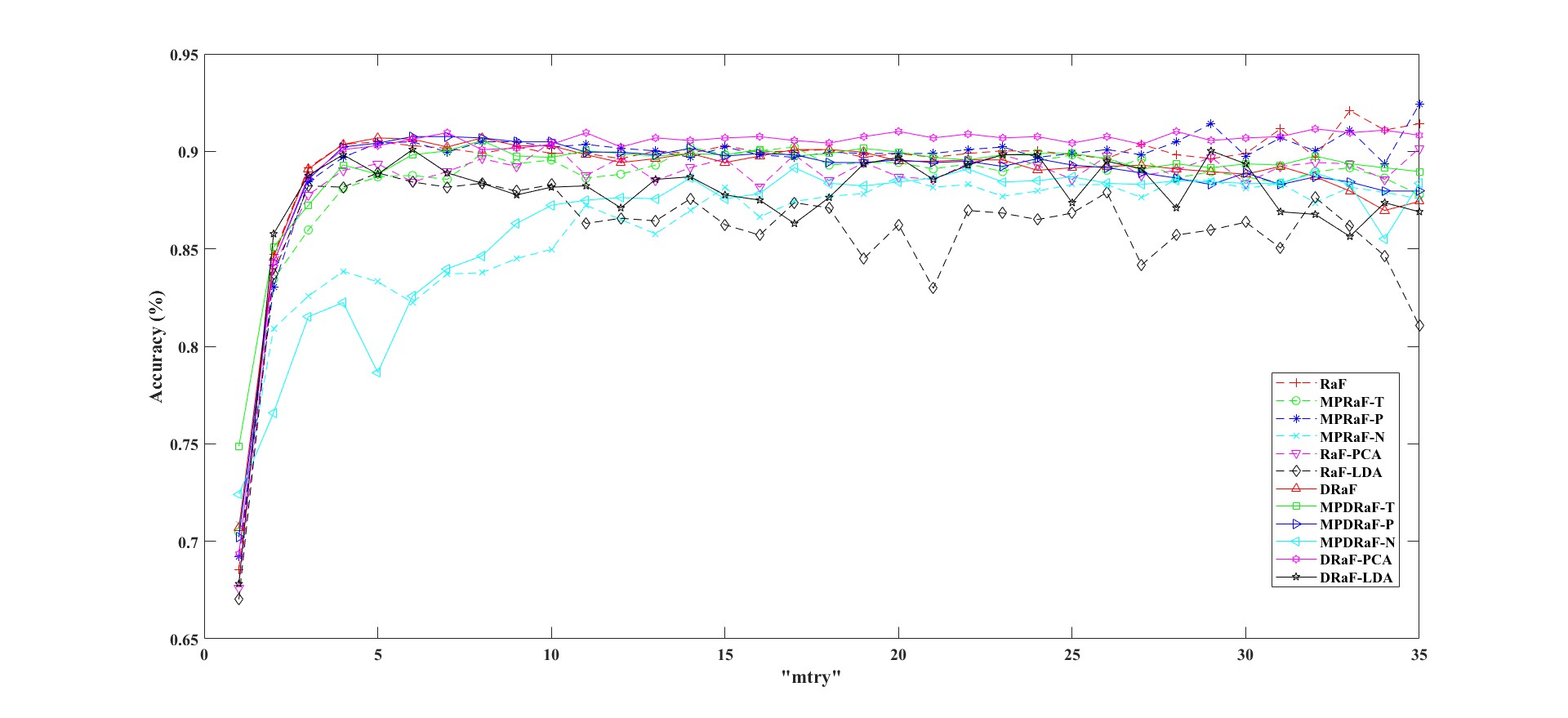}} 
\end{figure}

The parameter \textit{``mtry"} denotes the  number of candidate features to be evaluated at each non-leaf node. In a given problem, the smaller \textit{``mtry"} results in stronger randomization among the trees and weaker dependency of their structures on the output. However, if the \textit{``mtry"} is small, the random  subset of features selected at a given node may fail to get the geometry of the data points. To see the effect of \textit{``mtry"} parameter, we varied it to different values on the  datasets given in Figure \ref{fig:effect of mtry}.
From the  Figure \ref{fig:effect of mtry}, it is clear that at very low values of \textit{``mtry"}, the performance is lower. However, as the size of the \textit{``mtry"} parameter increases, the performance starts increasing and becomes stable very quickly. Setting \textit{``mtry"} to $round(\sqrt{n})$ leads to satisfactory performance.

\subsection{Effect of \textit{``minleaf"} parameter}
In the ensembles of decision tree \textit{``minleaf"} denotes the  maximum number of data samples to be placed in an impure node. In general, smaller trees are generated with higher minleaf which results in higher bias and lower variance. \citet{zhang2008rotboost} suggested that performance ensembles of decision tree are robust to this parameter while as  \citet{lin2006random} suggested that its optimal value varies in different situations. To analyse the effect of this parameter, we evaluated the effect of \textit{``minleaf"} parameter with its value varying from $1$ to $3$ on $120$ datasets (leaving miniboone dataset as it took huge time to compute for all these parameters). The average rank of each model across different parameters corresponding to each model are given in Table \ref{tab:effect_of_minleaf}. 
With $N=120$, $K=3$ (as minleaf=1,2,3), $F_F(2,238)=3.03$. Significant difference exist among the different performances based on the minleaf value of the model if $F_F>3.03$ (Table \ref{tab:effect_of_minleaf}).
From the given table, it is clear that significant difference exists among the performances of the all the models (except DRaF-LDA) with different minleaf parameters. However, in most of the cases smaller minleaf parameter results in better performance. This study is in consensus with the observation that decision trees of an ensemble should grow as much as possible for better performance.

    \begin{longtable}{lcccc}
    \caption{Average rank of the classification models with different \textit{minleaf} parameters.}
    \label{tab:effect_of_minleaf}\\
    \endfirsthead
\multicolumn{5}{c}%
{{\bfseries \tablename\ \thetable{} -- continued from previous page}} \\
\hline
 Method& $minleaf=1$&$minleaf=2$&$minleaf=3$&$F_F$\\
 \hline 
\endhead
\hline \multicolumn{5}{|r|}{{Continued on next page}} \\ \hline
\endfoot
\endlastfoot

    \hline
    Method& $minleaf=1$&$minleaf=2$&$minleaf=3$&$F_F$\\
    \hline
         RaF&$1.87$&$1.87$&$2.26$&$6.3555$\\
MPRaF-T&$1.79$&$1.93$&$2.29$&$11.2753$\\
MPRaF-P&$1.66$&$2.11$&$2.23$&$11.8124$\\
MPRaF-N&$1.73$&$1.95$&$2.32$&$11.6113$\\
RaF-PCA&$1.73$&$2.01$&$2.27$&$12.1945$\\
RaF-LDA&$1.77$&$2.01$&$2.22$&$6.3555$\\
DRaF&$1.65$&$2.08$&$2.27$&$13.3546$\\
MPDRaF-T$^*$&$1.87$&$1.94$&$2.19$&$3.4658$\\
MPDRaF-P$^*$&$1.74$&$1.98$&$2.28$&$9.3988$\\
MPDRaF-N$^*$&$1.76$&$1.93$&$2.3$&$7.0927$\\
DRaF-PCA$^*$&$1.68$&$2$&$2.32$&$13.5758$\\
DRaF-LDA$^*$&$1.8$&$2.08$&$2.11$&$1.111$\\
  \hline
  
    \end{longtable}


\subsection{Average Number of Nodes}
\begin{figure}
    \centering
    \includegraphics[width=\textwidth]{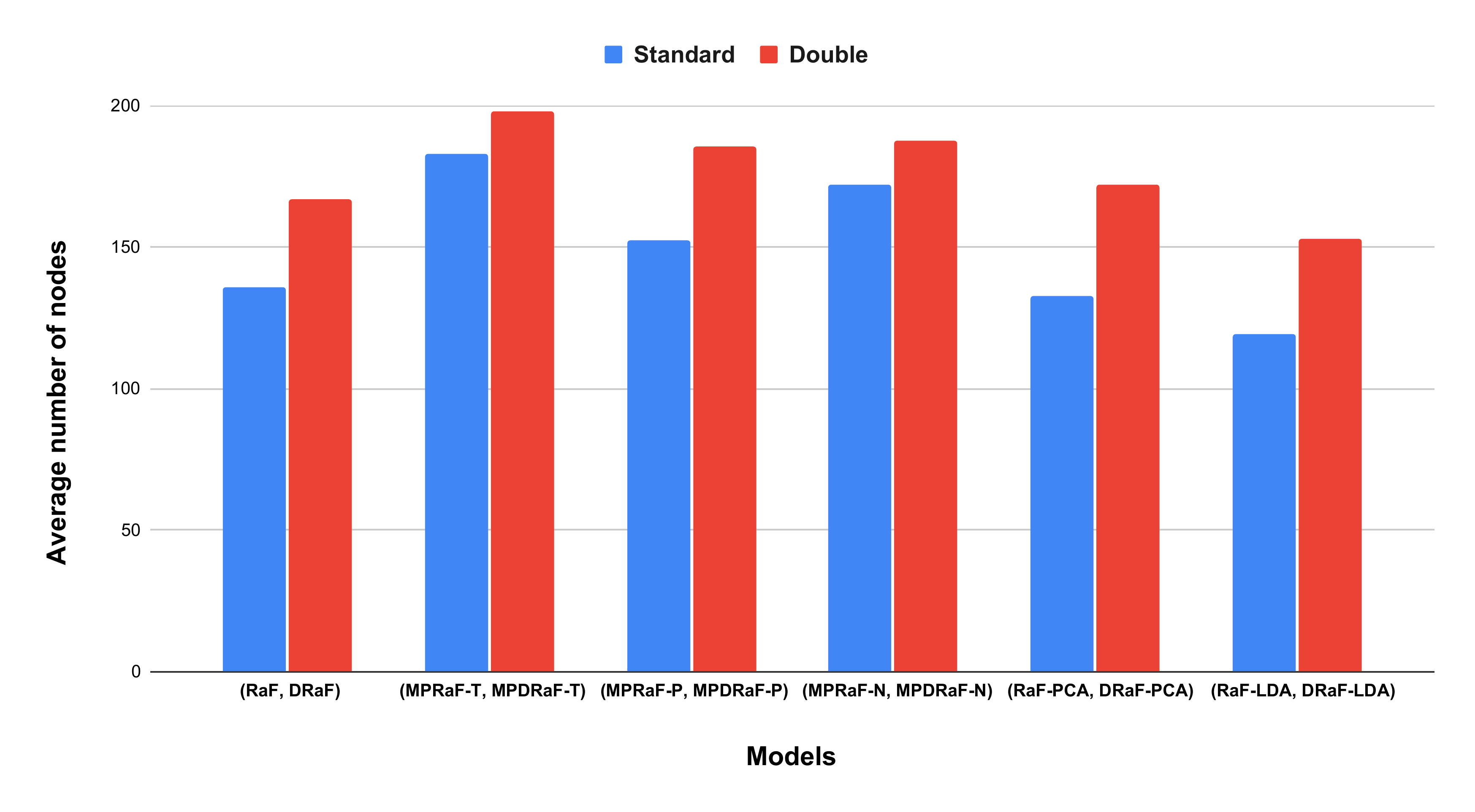}
    \caption{Mean Node Analysis}
    \label{fig:bar chart Mean Node Analysis}
\end{figure}

As seen in the above section that smaller minleaf results in better performance, hence, the performance of the models can be increased if there is a way to generate the bigger trees \cite{han2020double}. Thus, greater the size of the tree better the performance is. Here, we analyse the size of the tree via  number of nodes. Average number of nodes denote that the average number of  nodes in an ensemble.
Table \ref{table:Average number of nodes} gives the average of the nodes present in different ensembles of the classification models. From Figure \ref{fig:bar chart Mean Node Analysis} represents the average of mean nodes in different classification models.
Figure \ref{fig:bar chart Mean Node Analysis}, it is clear that double variants of the random forest have higher number of nodes compared to the standard variants  of the random forest. Hence, the proposed variants of the double random forest show better performance due to larger size of the trees. 

\begin{landscape}
\begin{footnotesize}
\begin{longtable}[t]{|p{0.18\textwidth}|p{0.07\textwidth}|p{0.08\textwidth}|p{0.08\textwidth}|p{0.08\textwidth}|p{0.1\textwidth}|p{0.1\textwidth}|c|c|c|c|c|c|p{0.1\textwidth}|}
\caption[this is]{Average number of nodes
in RaF \cite{breiman2001random}, MPRaF-T \cite{zhang2014oblique}, MPRaF-P \cite{zhang2014oblique}, MPRaF-N \cite{zhang2014oblique}, RaF-PCA \cite{zhang2014random}, RaF-LDA \cite{zhang2014random}, DRaF \cite{han2020double}, MPDRaF-T, MPDRaF-P, MPDRaF-N, DRaF-PCA AND DRaF-LDA classification models } \label{table:Average number of nodes} \\
\hline
Datasets &RaF &MPRaF-T&MPRaF-P&MPRaF-N&RaF-PCA&RaF-LDA&DRaF&MPDRaF-T$^*$&MPDRaF-P$^*$&MPDRaF-N$^*$&DRaF-PCA$^*$&DRaF-LDA$^*$\\
\hline 
\endfirsthead
\multicolumn{13}{c}%
{{\bfseries \tablename\ \thetable{} -- continued from previous page}} \\
\hline
Datasets &RaF &MPRaF-T&MPRaF-P&MPRaF-N&RaF-PCA&RaF-LDA&DRaF&MPDRaF-T$^*$&MPDRaF-P$^*$&MPDRaF-N$^*$&DRaF-PCA$^*$&DRaF-LDA$^*$\\
 \hline 
\endhead
\hline \multicolumn{13}{|r|}{{Continued on next page}} \\ \hline
\endfoot
\endlastfoot

\hline
       abalone&$349.81$&$481.93$&$392.75$&$372.79$&$304.43$&$296.08$&$470.07$&$447.08$&$491.77$&$324.92$&$444.69$&$439.34$\\
acute-inflammation&$4.9$&$5.04$&$4.97$&$4.74$&$4.46$&$4.44$&$5.05$&$4.99$&$5.18$&$4.72$&$4.32$&$4.54$\\
acute-nephritis&$3.89$&$4.46$&$4.03$&$4.09$&$3.57$&$3.66$&$4.01$&$4.48$&$4.3$&$4.27$&$3.52$&$3.77$\\
adult&$1731.66$&$2113.55$&$1843.9$&$1846.68$&$1480.15$&$1489.83$&$2132.73$&$2456.92$&$2321.47$&$2021.68$&$2016.47$&$1965.98$\\
annealing&$34.1$&$55.84$&$34.65$&$46.51$&$32.62$&$34.05$&$40.54$&$68.02$&$40.73$&$54.3$&$38.96$&$40.03$\\
arrhythmia&$36.6$&$63.65$&$36.12$&$63.92$&$34.81$&$35.39$&$43.64$&$58.73$&$43.54$&$58.12$&$46.24$&$47.53$\\
audiology-std&$17.26$&$23.98$&$17.22$&$23.18$&$16.63$&$17.89$&$14.43$&$26.65$&$13.98$&$23.34$&$15.89$&$16.14$\\
balance-scale&$38.45$&$38.18$&$37.65$&$34.35$&$32.13$&$30.73$&$35.96$&$36.71$&$36.53$&$30.1$&$33.8$&$32.89$\\
balloons&$2.1$&$1.84$&$2.13$&$1.78$&$2.23$&$2.33$&$1.25$&$1.28$&$1.38$&$1.32$&$1.92$&$1.86$\\
bank&$167.83$&$245.75$&$185.29$&$218.47$&$140.31$&$138.89$&$232.43$&$294.41$&$250.61$&$264.36$&$207.68$&$203.55$\\
blood&$44.22$&$39.85$&$47.36$&$44.86$&$41.17$&$40.84$&$54.81$&$41.74$&$54.41$&$46.77$&$56.34$&$56.92$\\
breast-cancer&$26.38$&$28.39$&$26.35$&$26.41$&$21.9$&$21.39$&$29.49$&$31.43$&$31.22$&$26.8$&$31.53$&$28.97$\\
breast-cancer-wisc&$14.94$&$17.11$&$14.83$&$17.75$&$11.36$&$10.34$&$19.77$&$19.92$&$18.68$&$22.86$&$16.07$&$15.21$\\
breast-cancer-wisc-diag&$11.41$&$14.47$&$11.79$&$18.57$&$11.87$&$8.36$&$16.65$&$18.06$&$15.71$&$23.04$&$17.41$&$11.69$\\
breast-cancer-wisc-prog&$12.66$&$17.26$&$13.81$&$19.59$&$11.65$&$9.77$&$19.42$&$21.51$&$20.14$&$23.98$&$19.2$&$17.14$\\
breast-tissue&$10.22$&$13.94$&$11.58$&$12.88$&$10.03$&$9.45$&$12.42$&$11.86$&$13.4$&$9.65$&$12.85$&$12.45$\\
car&$52.05$&$53.99$&$59.95$&$38.74$&$56.29$&$51.94$&$48.68$&$51.09$&$57.2$&$32.38$&$55.22$&$51.35$\\
cardiotocography-10clases&$134.08$&$242.91$&$144.6$&$253.69$&$134.46$&$113.43$&$161.43$&$256.29$&$174.09$&$270.15$&$176.92$&$148.62$\\
cardiotocography-3clases&$63.94$&$110.23$&$71.05$&$122.75$&$60.33$&$54.43$&$80.75$&$127.33$&$90.16$&$144.95$&$85.3$&$75.46$\\
chess-krvk&$922.92$&$1341.55$&$1138.44$&$651.32$&$1250.65$&$1226.39$&$720.29$&$897.58$&$876.76$&$390.42$&$1014.06$&$982.35$\\
chess-krvkp&$91.09$&$158.26$&$96.72$&$130.73$&$94.32$&$90.97$&$98.88$&$190.97$&$106.51$&$149.3$&$110.76$&$102.34$\\
congressional-voting&$9.03$&$11.01$&$8.65$&$8.95$&$8.6$&$8.12$&$6.46$&$13.07$&$6.49$&$10.67$&$8.65$&$8.58$\\
conn-bench-sonar-mines-rocks&$11.95$&$17.62$&$12.94$&$19.71$&$11.59$&$8.2$&$19.3$&$21.92$&$21.08$&$24.68$&$19.88$&$15.11$\\
conn-bench-vowel-deterding&$61.24$&$92.06$&$71.79$&$94.91$&$58.61$&$52.83$&$77.37$&$103.21$&$89.43$&$94.97$&$76.15$&$67.79$\\
connect-4&$1967.07$&$2058.1$&$1891.47$&$1183.77$&$2060.35$&$2069.17$&$2042.64$&$2231.71$&$1938.48$&$1325.44$&$2340.46$&$2169.71$\\
contrac&$132.21$&$143.02$&$134.2$&$112.51$&$122.42$&$118.87$&$131.37$&$124.59$&$137.45$&$94$&$144.96$&$138.52$\\
credit-approval&$38.91$&$57.35$&$43.62$&$52.44$&$34.06$&$32.43$&$52.69$&$67.08$&$56.72$&$61.46$&$51.09$&$47.16$\\
cylinder-bands&$41.29$&$57.62$&$42.98$&$58.21$&$32.63$&$30.92$&$59.68$&$63.21$&$63.02$&$63.14$&$52.22$&$51.39$\\
dermatology&$16.57$&$21.36$&$16.72$&$22.93$&$15.93$&$15.07$&$18.75$&$25.76$&$18.88$&$26.48$&$18.59$&$19.01$\\
echocardiogram&$9.39$&$12.91$&$10.54$&$13.51$&$8.47$&$8.16$&$12.5$&$13.4$&$13.18$&$13.64$&$12.37$&$11.55$\\
ecoli&$21.06$&$29.67$&$22.12$&$29.22$&$19.55$&$19.53$&$24.27$&$30.18$&$25.9$&$29.94$&$24.87$&$24.44$\\
energy-y1&$19.81$&$28.57$&$21.12$&$23.27$&$21.05$&$19.8$&$20.84$&$28.92$&$22.49$&$22.94$&$22.88$&$20.51$\\
energy-y2&$20.16$&$27.39$&$21.02$&$21.62$&$19.94$&$18.93$&$19.06$&$26.41$&$19.82$&$19.59$&$20.55$&$20.12$\\
fertility&$6.59$&$7.89$&$6.98$&$7.71$&$5.3$&$4.98$&$7.55$&$8.45$&$7.9$&$8.06$&$6.39$&$6.25$\\
flags&$22.14$&$29.94$&$23.32$&$30.2$&$19.74$&$19.1$&$20.38$&$24.02$&$20.65$&$23.11$&$22.19$&$22.11$\\
glass&$19.94$&$28.3$&$20.21$&$27.43$&$18.29$&$17.5$&$23.35$&$26.87$&$24.81$&$24.85$&$24.46$&$24.13$\\
haberman-survival&$25.08$&$30.43$&$27.5$&$30.54$&$22.39$&$22.72$&$31.43$&$30.7$&$29.78$&$29.58$&$31.97$&$31.9$\\
hayes-roth&$9.48$&$11.65$&$10.47$&$11.06$&$9.27$&$9.57$&$9.25$&$12.94$&$10.4$&$12.14$&$9.38$&$10.1$\\
heart-cleveland&$29.28$&$38.23$&$30.57$&$38.94$&$24.89$&$23.78$&$31.77$&$30.35$&$31.28$&$31.75$&$32.14$&$31.8$\\
heart-hungarian&$19.39$&$24.65$&$20.9$&$22.62$&$17.24$&$16.58$&$23.73$&$27.59$&$25.09$&$23.16$&$24.27$&$22.77$\\
heart-switzerland&$16.19$&$18.03$&$15.94$&$16.19$&$13.97$&$14.5$&$13.85$&$11.5$&$14.31$&$10.72$&$15.2$&$15$\\
heart-va&$24.8$&$29.49$&$24.4$&$25.87$&$22.4$&$22.69$&$21.02$&$17.98$&$21.08$&$16.34$&$25.44$&$25.66$\\
hepatitis&$10.18$&$13.48$&$10.88$&$12.77$&$8.45$&$7.74$&$13.07$&$14.49$&$13.59$&$13.81$&$11.91$&$11.54$\\
hill-valley&$66.19$&$35.12$&$78.26$&$66.38$&$44.66$&$33.52$&$107.02$&$48.96$&$113.11$&$81.06$&$71.38$&$57.43$\\
horse-colic&$24.01$&$35.35$&$26.48$&$36.79$&$21.26$&$18.99$&$30.83$&$42.35$&$35.99$&$43.91$&$32.1$&$29.08$\\
ilpd-indian-liver&$42.55$&$62.23$&$52.69$&$59.59$&$37.81$&$36.36$&$64.22$&$66.84$&$70.02$&$61.38$&$60.39$&$60.05$\\
image-segmentation&$15.59$&$27.32$&$18.32$&$28.83$&$15.1$&$13.31$&$18.55$&$29.1$&$22.34$&$31.26$&$19.57$&$17.15$\\
ionosphere&$13.52$&$21.57$&$16.08$&$23.27$&$12.08$&$10.21$&$18.44$&$27.29$&$22.89$&$29.9$&$18.13$&$15.86$\\
iris&$5.29$&$7.04$&$5.75$&$6.28$&$5.98$&$4.31$&$5.84$&$6.99$&$6.15$&$6.31$&$7.45$&$5.29$\\
led-display&$12.41$&$13.45$&$12.54$&$12.22$&$13.08$&$13.42$&$10.19$&$11.76$&$9.96$&$9.72$&$11.3$&$10.27$\\
lenses&$2.8$&$2.61$&$2.75$&$2.78$&$2.78$&$2.77$&$2.15$&$1.85$&$2.01$&$2.12$&$2.19$&$2.29$\\
letter&$1190.77$&$1941.31$&$1545.76$&$2019.17$&$1105.98$&$930.41$&$1277.75$&$1958.65$&$1592.73$&$1954.44$&$1307.54$&$1096.94$\\
libras&$35.05$&$52.7$&$41.16$&$56.74$&$33.61$&$24.04$&$46.59$&$58.94$&$50.94$&$60.77$&$45.32$&$35.67$\\
low-res-spect&$21.64$&$30.63$&$24.64$&$41.84$&$22.72$&$15.78$&$29.46$&$37.97$&$32.4$&$54.2$&$29.92$&$22.45$\\
lung-cancer&$4.49$&$4.47$&$4.37$&$4.87$&$4.04$&$3.63$&$4.23$&$3.11$&$4.42$&$3.11$&$4.3$&$4.2$\\
lymphography&$13.44$&$16.54$&$12.94$&$16.01$&$10.45$&$10.4$&$15.29$&$17.56$&$15.17$&$16.81$&$14.91$&$13.9$\\
magic&$778.89$&$1291.33$&$1027.86$&$1307.15$&$753.48$&$712.13$&$1111.39$&$1482.07$&$1405$&$1449.84$&$1107.32$&$1065.07$\\
mammographic&$44.23$&$41.93$&$46.28$&$37.9$&$38.79$&$40.24$&$55.53$&$50.51$&$55.49$&$44.05$&$51.62$&$52.76$\\
miniboone&$2878.18$&$3186$&$3325.18$&$3415.87$&$2829.45$&$2159.2$&$4173.28$&$3917.6$&$4949.34$&$4402.77$&$4223.4$&$3380.49$\\
molec-biol-promoter&$8.25$&$11.47$&$9.13$&$11.27$&$7.16$&$5.22$&$11.85$&$12.52$&$12.61$&$12.61$&$12.22$&$9.04$\\
molec-biol-splice&$176.93$&$313.37$&$203.18$&$340.19$&$189.1$&$121.92$&$222.68$&$344.12$&$267.96$&$367.3$&$270.48$&$185.86$\\
monks-1&$13.76$&$14.75$&$14.34$&$13.17$&$13.46$&$13.17$&$13.82$&$14.26$&$14.8$&$12.42$&$14.68$&$14.39$\\
monks-2&$18.58$&$19.18$&$19.15$&$17.15$&$19.37$&$20.08$&$17.63$&$15.94$&$17.98$&$13.79$&$21.23$&$20.87$\\
monks-3&$11.2$&$12.73$&$12.65$&$11.99$&$10.42$&$10.54$&$12.02$&$12.08$&$12.44$&$11.3$&$11.67$&$11.34$\\
mushroom&$22.75$&$40.82$&$28.52$&$40.07$&$25.59$&$22.62$&$22.69$&$41.7$&$29.31$&$40.72$&$25.97$&$22.32$\\
musk-1&$24.84$&$38.13$&$27.21$&$42.76$&$23.55$&$14.11$&$38.03$&$49.24$&$42.81$&$56.82$&$39.26$&$25.85$\\
musk-2&$122.19$&$209.28$&$154.56$&$269.33$&$131.13$&$80.8$&$155.73$&$272.22$&$218.71$&$358.09$&$185.13$&$119.86$\\
nursery&$251.01$&$304.95$&$292.95$&$227.6$&$284.02$&$229$&$238.47$&$317.62$&$291.65$&$207.74$&$284.79$&$238.4$\\
OM\_nucleus\_4d&$60.42$&$82$&$67.16$&$77.37$&$49.74$&$41.21$&$90.32$&$96.09$&$97.68$&$97.48$&$78.41$&$65.48$\\
OM\_states\_2f&$28.9$&$43.9$&$34.34$&$50.86$&$29.16$&$22.92$&$41.53$&$54.12$&$47.79$&$63.07$&$41.68$&$34.74$\\
OT\_nucleus\_2f&$54.49$&$81$&$67.18$&$87.08$&$49.45$&$43.09$&$81.65$&$97.52$&$96.57$&$105.17$&$78.13$&$69.05$\\
OT\_states\_5b&$35.15$&$48.35$&$36.61$&$60.54$&$32.36$&$24.35$&$50.86$&$61.42$&$54.25$&$80.15$&$47.17$&$36.41$\\
optical&$210.94$&$383.82$&$219.43$&$467.41$&$204.48$&$183.58$&$244.67$&$454.88$&$253.86$&$562.79$&$246.3$&$218.09$\\
ozone&$28.65$&$41.55$&$31.06$&$52.34$&$27.68$&$20.37$&$42.9$&$56.45$&$45.01$&$74.71$&$40.87$&$31.9$\\
page-blocks&$63.65$&$91.79$&$72.87$&$85.22$&$59.51$&$59$&$78.39$&$102.96$&$92.8$&$89.51$&$77.12$&$77.25$\\
parkinsons&$8.53$&$12.85$&$10.47$&$13.85$&$8.53$&$7.28$&$12.53$&$15.65$&$14.38$&$16.36$&$12.63$&$10.9$\\
pendigits&$210.57$&$320.17$&$237.25$&$385.79$&$200.55$&$153.25$&$254.88$&$371.52$&$279.76$&$434.7$&$245.63$&$188.28$\\
pima&$49.77$&$73.48$&$61.52$&$77.61$&$45.37$&$43.15$&$71.98$&$79.12$&$81.43$&$81.48$&$71.59$&$69$\\
pittsburg-bridges-MATERIAL&$7.64$&$8.8$&$7.63$&$8.55$&$6.54$&$6.35$&$8.78$&$9.04$&$8.5$&$8.33$&$7.39$&$7.36$\\
pittsburg-bridges-REL-L&$11.41$&$12.56$&$11.42$&$12.48$&$9.7$&$9.55$&$11.93$&$12.62$&$12.23$&$11.47$&$11.32$&$10.99$\\
pittsburg-bridges-SPAN&$10.1$&$11.39$&$10.42$&$11.4$&$8.78$&$8.55$&$10.38$&$10.78$&$10.49$&$9.89$&$9.76$&$9.72$\\
pittsburg-bridges-T-OR-D&$6.56$&$7.48$&$6.65$&$7.78$&$5.28$&$4.93$&$7.27$&$8.55$&$7.66$&$8.49$&$6.45$&$5.95$\\
pittsburg-bridges-TYPE&$12.21$&$14.65$&$12.47$&$14.11$&$11.06$&$11.04$&$11.24$&$11.1$&$11.46$&$11.3$&$10.85$&$10.98$\\
planning&$14.76$&$21.19$&$18.04$&$22.72$&$14.07$&$13.81$&$22.81$&$22.24$&$24.48$&$22.64$&$22.64$&$22.4$\\
plant-margin&$185.17$&$278.12$&$226.79$&$284.17$&$177.44$&$162.35$&$208.88$&$237.69$&$224.82$&$215.9$&$214.61$&$198.11$\\
plant-shape&$179.46$&$269.87$&$225.2$&$266.63$&$172.3$&$151.38$&$211.7$&$223.83$&$224.9$&$208.34$&$211.32$&$185.95$\\
plant-texture&$183.56$&$280.18$&$229.79$&$288.43$&$185.23$&$174.11$&$204.49$&$239.06$&$224.63$&$228.76$&$213.6$&$203.7$\\
post-operative&$9.47$&$10.15$&$9.36$&$9.94$&$8.33$&$8.2$&$8.78$&$9.8$&$9.02$&$8.52$&$9.1$&$8.89$\\
primary-tumor&$30.61$&$35.14$&$30.93$&$29.84$&$29.14$&$27.48$&$28.67$&$28.18$&$27.73$&$25.73$&$26.43$&$23.9$\\
ringnorm&$192.09$&$94.06$&$78.06$&$97.7$&$184.54$&$177.42$&$278.44$&$94.96$&$81.93$&$106.52$&$276.47$&$270.66$\\
seeds&$7.75$&$9.65$&$8.57$&$9.61$&$7.46$&$5.71$&$10.24$&$10.99$&$10.03$&$10.33$&$9.25$&$7.65$\\
semeion&$133.92$&$200.29$&$140.83$&$211.13$&$112.95$&$74.38$&$134.27$&$205.83$&$142.29$&$217.42$&$129.69$&$95.45$\\
soybean&$34.87$&$43.67$&$34.94$&$42.07$&$31.01$&$30.76$&$38.45$&$48.22$&$38.59$&$46.51$&$35.89$&$34.99$\\
spambase&$145.86$&$266.25$&$155.26$&$264.33$&$126.5$&$128.33$&$185.34$&$359.57$&$195.29$&$330.51$&$172.87$&$172.18$\\
spect&$8.87$&$10.66$&$9.01$&$10.16$&$7.56$&$7.34$&$7.92$&$11.14$&$8.36$&$9.83$&$9.03$&$8.89$\\
spectf&$7.22$&$9.79$&$7.9$&$8.53$&$6.88$&$5.17$&$10.53$&$12.95$&$11.62$&$11.17$&$11.03$&$8.53$\\
statlog-australian-credit&$61.65$&$84.09$&$67.34$&$80.9$&$51.56$&$50.41$&$88.45$&$86.61$&$93.81$&$81.66$&$82.81$&$81.69$\\
statlog-german-credit&$83.54$&$104.95$&$86.89$&$98.17$&$62.34$&$61.28$&$114.18$&$115.94$&$116.87$&$111.89$&$96.46$&$93.76$\\
statlog-heart&$19.09$&$23.63$&$19.61$&$24.06$&$15.01$&$13.64$&$25.98$&$27$&$25.5$&$27.88$&$22.85$&$21.07$\\
statlog-image&$53.56$&$104.54$&$66.08$&$110.84$&$55.86$&$44.66$&$64.94$&$122.42$&$82.5$&$126.87$&$70.88$&$57.08$\\
statlog-landsat&$212.51$&$282.05$&$215.65$&$348.32$&$183.51$&$147.53$&$279.69$&$349.92$&$291.84$&$441.74$&$255.52$&$214.28$\\
statlog-shuttle&$43.13$&$85.75$&$50.42$&$88.78$&$50.5$&$46.36$&$46.61$&$96.3$&$55.25$&$92.36$&$56.93$&$53.09$\\
statlog-vehicle&$62.92$&$86.28$&$69.01$&$86.57$&$54.83$&$48.37$&$84.86$&$88.85$&$89.87$&$86.25$&$80.76$&$72.17$\\
steel-plates&$137.59$&$219.51$&$151.26$&$223.54$&$127.45$&$116.3$&$174.26$&$212$&$191.89$&$216.55$&$172.86$&$163.19$\\
synthetic-control&$22.45$&$30.57$&$24.67$&$47.6$&$22.3$&$14.81$&$31.6$&$39.99$&$33.12$&$68.44$&$31.62$&$23.2$\\
teaching&$16.53$&$15.55$&$15.79$&$13.96$&$14.67$&$15.2$&$17.06$&$13.03$&$16.26$&$10.53$&$16.72$&$16.82$\\
thyroid&$35.33$&$97.28$&$38.4$&$86.65$&$43.65$&$40.25$&$40.6$&$125.71$&$41.32$&$102.93$&$53.44$&$49.49$\\
tic-tac-toe&$63.59$&$74.08$&$68.68$&$70.4$&$53.15$&$49.43$&$69.82$&$89.25$&$79.65$&$78.96$&$68.94$&$60.88$\\
titanic&$2.11$&$1.79$&$1.96$&$1.22$&$1.89$&$1.92$&$1.78$&$1.57$&$1.67$&$1.02$&$1.66$&$1.58$\\
trains&$1.86$&$1.83$&$1.82$&$1.84$&$1.85$&$1.87$&$1.86$&$1.33$&$1.83$&$1.29$&$1.62$&$1.79$\\
twonorm&$228.12$&$218.97$&$174.37$&$246.81$&$148.81$&$117.17$&$338.36$&$276.83$&$242.89$&$304.57$&$213.91$&$179.02$\\
vertebral-column-2clases&$16.76$&$24.45$&$21.93$&$22.62$&$17.19$&$16.12$&$23.89$&$26.32$&$26.1$&$24.54$&$24.36$&$23.27$\\
vertebral-column-3clases&$17.79$&$27.44$&$22.68$&$23.81$&$19.01$&$17.86$&$24.66$&$27.57$&$27$&$25.11$&$26.38$&$24.15$\\
wall-following&$75.12$&$447.26$&$206.66$&$457$&$200.73$&$172.75$&$87.98$&$525.26$&$292.24$&$542.77$&$260.06$&$226.96$\\
waveform&$235.29$&$341.47$&$263.81$&$393.89$&$218.79$&$171.81$&$345.63$&$422.09$&$393.9$&$497.19$&$334.64$&$274.58$\\
waveform-noise&$248.63$&$394.91$&$283.88$&$450.01$&$252.12$&$178.42$&$364.23$&$477.08$&$430.04$&$541.41$&$371.84$&$285.08$\\
wine&$6.59$&$8.7$&$7.12$&$11.46$&$6.78$&$5.45$&$8.63$&$10.95$&$8.85$&$12.68$&$9.05$&$7.46$\\
wine-quality-red&$145.57$&$216.6$&$160.1$&$214.39$&$129.81$&$128.37$&$187.58$&$205.59$&$199.63$&$196.07$&$182.78$&$180.84$\\
wine-quality-white&$469.5$&$680.6$&$513.65$&$660.92$&$417.41$&$409.4$&$596.95$&$643.95$&$635.97$&$596.16$&$573.59$&$569.44$\\
yeast&$154.13$&$205.48$&$159.04$&$180.27$&$139.59$&$140.98$&$173.4$&$158.29$&$176.88$&$145.17$&$175.34$&$176.23$\\
zoo&$7.05$&$8.55$&$7.22$&$8.02$&$6.7$&$6.63$&$7.21$&$9$&$7.41$&$8.29$&$7.29$&$7.32$\\
\hline
Average of Mean Nodes&$135.98$&$183.01$&$152.44$&$172.21$&$132.99$&$119.17$&$166.91$&$198.03$&$185.75$&$187.43$&$171.87$&$153.07$\\
\hline
\multicolumn{13}{l}{Here, OM denotes oocytes\_merluccius, OT denotes oocytes\_trisopterus.}
    \end{longtable}
    \end{footnotesize}
\end{landscape}

\section{Diversity Error Diagrams}
In this section, we analyze the existing baseline models and the proposed oblique and rotation double random forest in terms of ``diversity" among the individual decision tree classifiers and their classification accuracy or error. To visualise both the models in terms of these measures, visualization approach known as kappa-error diversity  diagrams are used \cite{margineantu1997pruning}. Kappa error diagrams use $2D$ plot for visualisation of individual accuracy and diversity of the members of the base learner. For $L$ number of base learners (here, decision trees) in an ensemble, a diagram is shown as a scatter plot of $L(L-1)/2$ points with each point corresponding to a pair of classifiers being analysed. The $x$-coordinate represents the diversity among the pair of base learners, also known as Kappa ($\kappa$) coefficient and the $y$-coordinate represents the average error of the pair of base learners. Kappa gives the level of agreement between the two base learners and while correcting for chance. For $T$ target labels of given dataset, $\kappa$ is defined on the $T \times T$  coincidence matrix $C$ of two classifiers. Each entry in the $c_{ij}$ represents the proportion of the testing data which one classifier predicted as $k^{th}$ class while the other base learner classifies it as the $j^{th}$ class. Kappa coefficient $\kappa$ represents the level of agreement between the two classifiers and is given as follows:
\begin{align}
    \kappa=\frac{p_r(a)-p_r(e)}{1-p_r(e)}
\end{align}
where $p_r(a)$ is the observed agreement between the two classifiers i.e. probability that both classifiers predicted the same label
and the $p_r(e)$ is the hypothetical probability of agreement by chance. Mathematically,
\begin{align}
    p_r(a)&=\sum_i c_{ii}, \\
    p_r(e)&=\sum_k\Bigg[ \Big(\sum_i m_{ki}\Big) \Big(\sum_j m_{jk}\Big) \Bigg].
 \end{align}

\begin{figure} 
\centering
\caption{Centroid of Kappa error diagrams on different datasets.}
\label{fig:Kappa_average_new}
    \subcaptionbox{ \label{fig:semeion}} { %
      \includegraphics[width=0.45\textwidth]{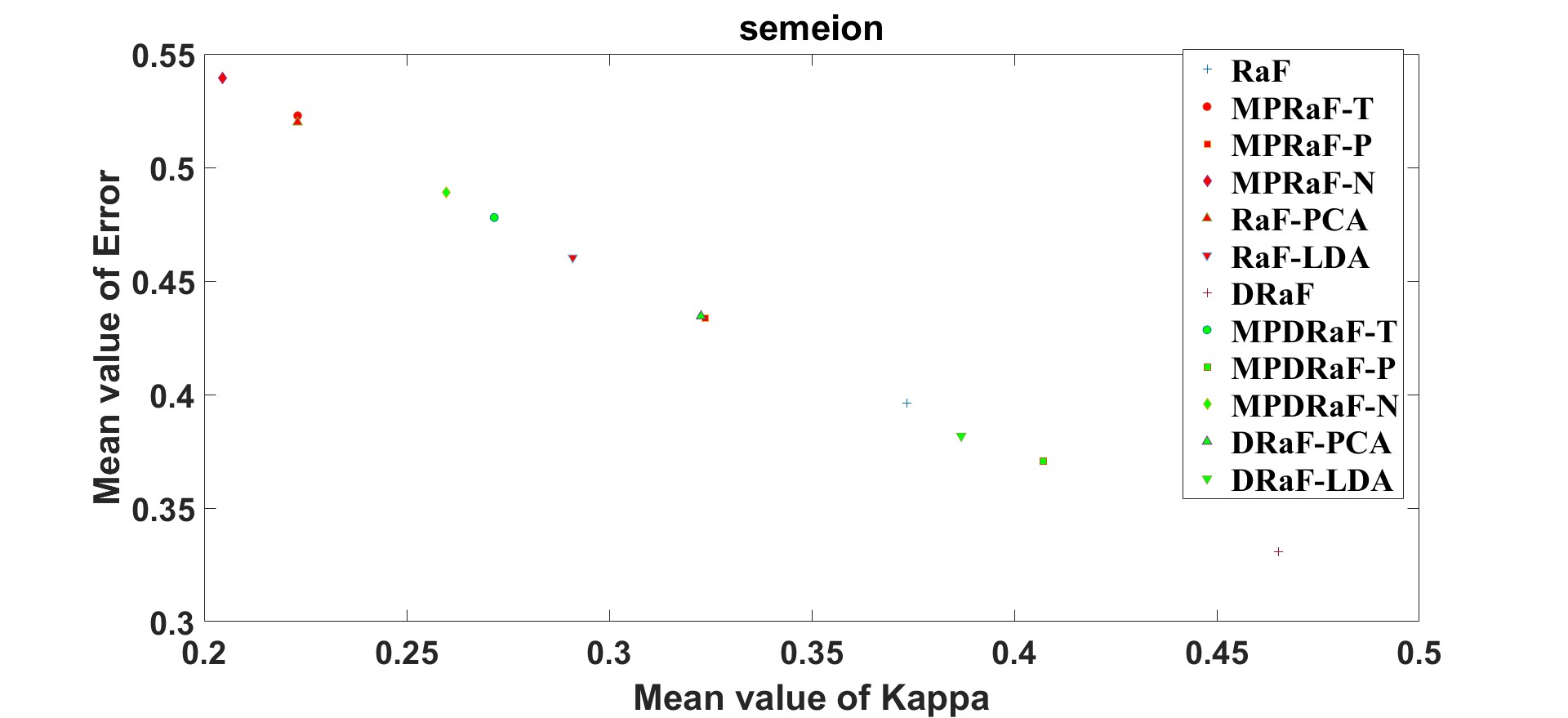}}
      \hfill
    \subcaptionbox{ \label{fig:oocytes_trisopterus_nucleus_2f}} { %
      \includegraphics[width=0.45\textwidth]{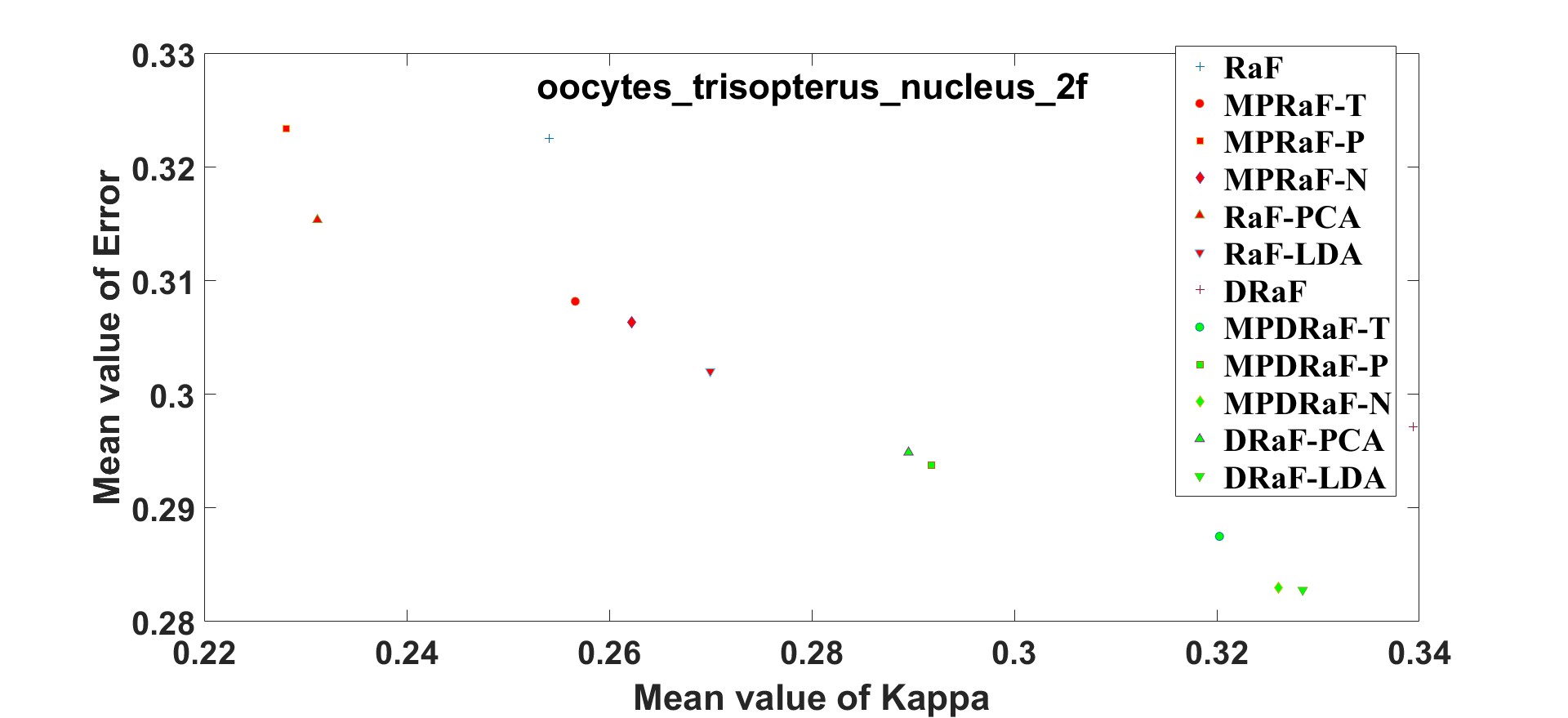}}
      \\
     \subcaptionbox{ \label{fig:oocytes_merluccius_nucleus_4d}} { %
      \includegraphics[width=0.45\textwidth]{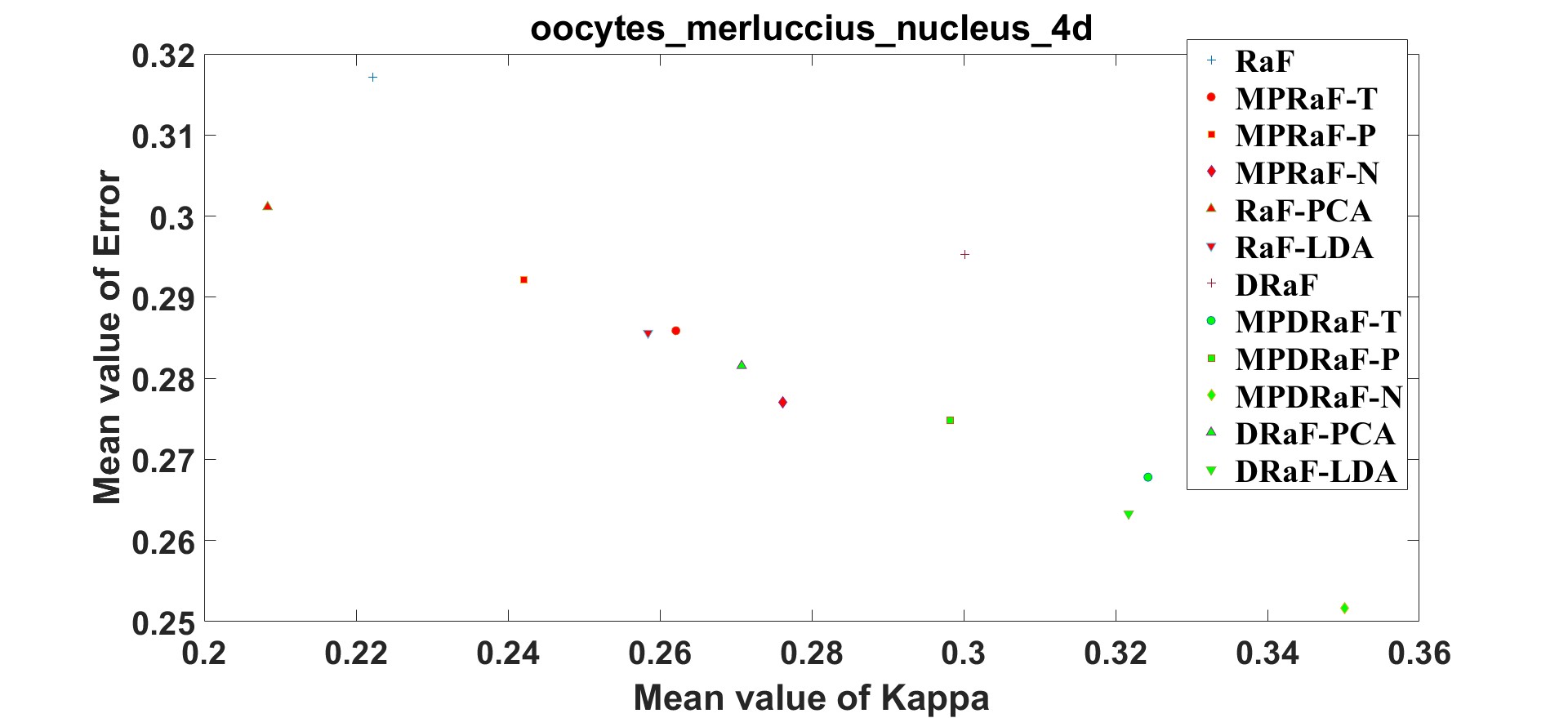}}
      \hfill
    \subcaptionbox{\label{fig:statlog-vehicle}} { %
      \includegraphics[width=0.45\textwidth]{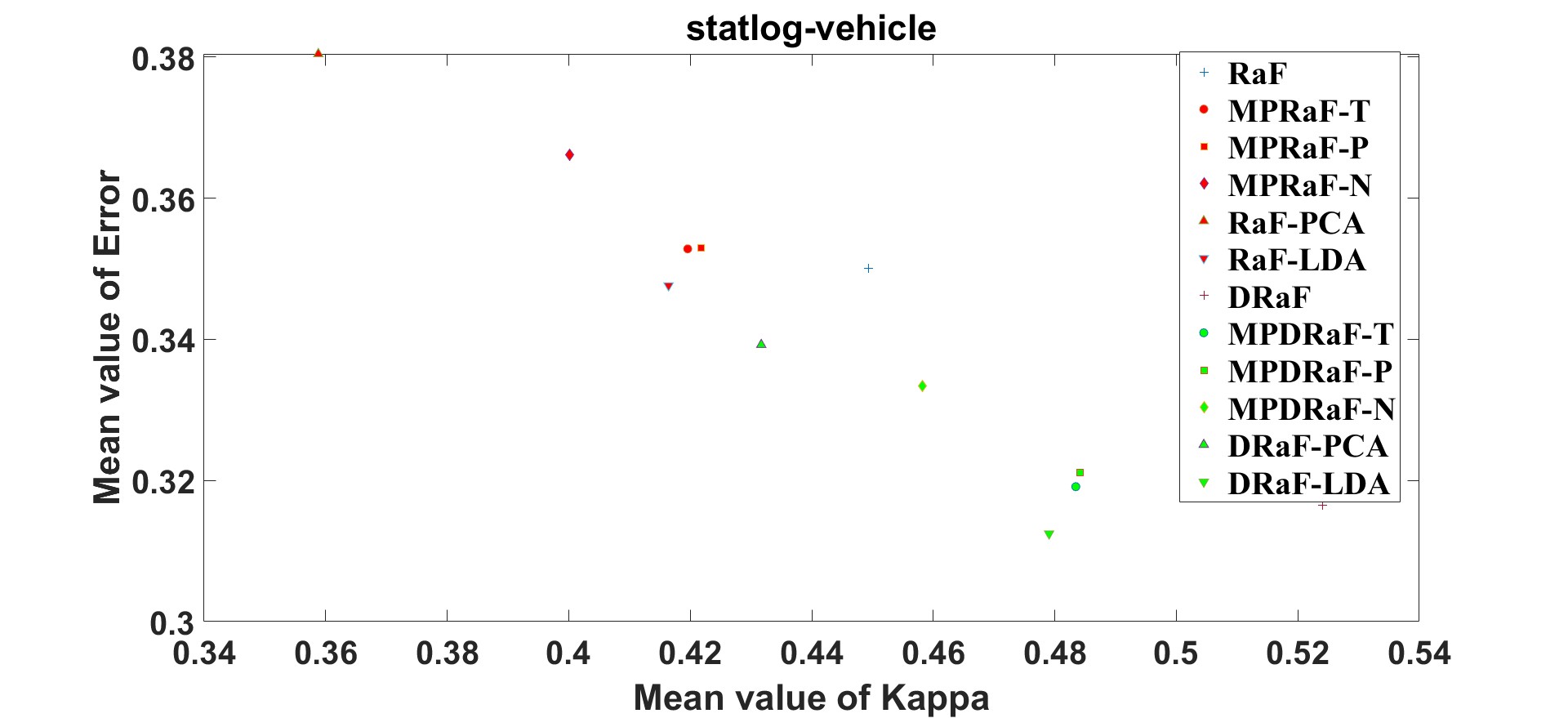}}
\end{figure}

If the two decision trees  are in complete agreement, then the kappa coefficient ($\kappa$) is $1$ and the two trees are identical. If the trees are independent, then the kappa coefficient ($\kappa$) is $0$. As mentioned above, we evaluate $L(L-1)/2$ pairs of kappa coefficients. Also, averaged error of the individual classifiers $E_{i,j}=(E_i+E_j)/2$. The smaller $\kappa$ value indicates better diversity or low correlation while as the smaller averaged error $E$ represents the more accurate or better strength classifier. The most desirable pair of classifiers is the one in the bottom left corner of Figure \ref{fig:Kappa_average_new}.  

Figure \ref{fig:Kappa_average_new} plots the kappa error diagram for some datasets. 
Figure $1,2,3$ and $4$ of the supplementary file shows the kappa error diagram for the semeion, oocytes\_trisopterus\_nucleus\_2f,  oocytes\_merluccius\_nucleus\_4d and statlog-vehicle datasets. The ensemble size is $50$, hence, $1225$ dots in each plot.
Figure $1(a)$  to Figure $1(l)$ of the supplementary file show the kappa error diagrams of the classification models on different datasets. All the classification models are trained on the training data samples and $\kappa$-error diagrams are plotted based on the performance of the classification models on the testing samples (in some diagrams the axis are adjusted for better view).

Figure \ref{fig:Kappa_average_new} represents the centroid of the scatter points for each classification model corresponding to the semeion, oocytes\_merluccius\_nucleus\_4d, oocytes\_trisopterus\_nucleus\_2f  and statlog-vehicle datasets. 
 From the given plots, different models of the random forest possess different characteristics. 
 Figure \ref{fig:semeion} plot shows that MPRaF-N is the most diverse classifier (least mean value of kappa) and DRaF is the most accurate classifier (least mean value of error). 
 However, DRaF-LDA ensemble classifiers possess the best overall generalization performance on this dataset. From the plot, one can see that the proposed DRaF-LDA have the better combination of diversity and error. Similarly in other datasets, the models with better combination results in better performance.
 
\section{Analysis of computational Complexity}
Here, we evaluate the  computational complexity of the classifiers. Without  assuming any structure of decision trees, we focus on the complexity involved at a given node. Let a given node receives $m$ number of samples with $n$ number of features.
In axis parallel splits, the optimum threshold is chosen based on some impurity criteria via ranking of each feature. Despite the complexity of the gini impurity, the complexity of the search involved in optimal split is   $O(nm~log~m)$ \cite{zhang2014oblique}. For MPSVM based oblique decision trees, the computational complexity of  generalized problem is $O(n^3)$ \cite{manwani2011geometric}. In  decision trees wherein the feature transformations (PCA and LDA) are used for projecting the input features, additional computational time is involved for calculating the projection matrix. The complexity of the PCA is $O(mn\times min(m,n)+n^3)$ \cite{kressner2004numerical} while as for LDA the complexity is $O(mn^2)$ \cite{chu2007map}. MPSVM based decision tree ensembles are faster as compared to the standard ensemble models. The reason is that in most of the cases, particularly for the nodes near the root, MPSVM method is faster compared to the exhaustive search. The training time of the proposed DRaF-PCA and DRaF-LDA is more as compared to the RaF-PCA and RaF-LDA, respectively, due to the reason that the bootstrapping at each non-leaf node of the proposed DRaF-PCA and DRaF-LDA leads to more number of unique samples to be sent down the tree resulting in more deeper decision trees. The average training time of each classification model is given in Table \ref{tab:average Friedman Rank}. The training time of the classification models on each dataset is given in Table 2 of the supplementary file.

\section{Bias Variance Analysis}
In this section, we discuss the bias-variance analysis of the ensemble models. Bias-variance analysis is the main reason for the success of ensemble models. The concept of bias-variance is well known in the regression problems for the squared loss functions \cite{geman1992neural}. However, this analysis is inappropriate as the labels of the classes are categorical. Thus, it is not feasible to transplant the decomposition of error in regression problems to classification problems. In classification problems, several studies have provided the ways to decompose the classification error into bias-variance terms \cite{kong1995error,friedman1997bias,james2003variance}. Each of these studies provide some insight into the models performance.

In this study, we consider $0-1$ loss function to analyse the performance of the models \cite{kohavi1996bias}. For details of bias-variance analysis via $0-1$ loss, refer to Section $5$ of the supplementary file. Let $D$ and $Y$ be spaces representing the input and output, respectively. Suppose $|D|$ represents the cardinality of $D$ and $|Y|$ represents the cardinality of $Y$. Also, let $d\in D$ and $y \in Y$ be the element its label respectively.
The conditional probability distribution of target $f$ is $P(Y_F=y_F|d)$ where $Y_F$ is the $Y$-valued random variable. Then for a single test data sample:
 \begin{align}
     E(C)=\sum_d P(D) [(bias_d)^2+\sigma_d^2+variance_d],
 \end{align}
 where
 \begin{align}
     (bias_d)^2&=\frac{1}{2}\sum_{y\in Y} [P(Y_F=y)-P(Y_H=y)]^2,\\
     variance_d&=\frac{1}{2}[1-\sum_{y\in Y} P(Y_H=y)^2],\\
     \sigma_d^2&=\frac{1}{2}[1-\sum_{y\in Y} P(Y_H=y)^2].
 \end{align}

Here, $(bias_d)^2$ and $variance_d$ are calculated are each model and for each dataset. $(bias_d)^2$ is abbreviated as $bias_d$. Theoretically, the error should be decomposed into squared bias, variance and noise (also known as irreducible error). However, given the real-world tasks wherein the true underlying probability distribution is unknown, estimation of noise is difficult task. In commonly used approach, the noise is generally aggregated into bias and variance or the only bias term as the noise in invariant across the learning models for a given task and hence not a significant factor for the comparative analysis of the algorithms. Table \ref{table:Bias_Varaince_Table} gives the bias-variance values for each model corresponding to the $121$  datasets. In most of the cases the double variant ensembles of decision trees have the best bias-variance values compared to the standard ensembles of the decision trees.

We evaluate the bias-variance of the classification models via statistical tests. In this test, the lower value of bias/variance gets lower rank and vice versa. The analysis of the results for bias and variance are given  in Table \ref{tab:bias_statitics} and Table \ref{tab:var_statistics}, respectively. From the given tables, it is clear that the double variants of the random forest achieve lower average rank compared to the standard variants  of random forest for both bias and variance performance. Hence, the proposed double variants of random forest  show better bias-variance results compared to the standard variants of the random forest. Moreover, the all the proposed variants of the double random forest are significantly better compared to the standard variants of the random forest.

\begin{landscape}
\small
\setlength{\tabcolsep}{2pt}

\begin{longtable}{@{\extracolsep{\fill}}|*{13}{c|}}
\caption[this is]{Bias variance analysis
of RaF \cite{breiman2001random}, MPRaF-T \cite{zhang2014oblique}, MPRaF-P \cite{zhang2014oblique}, MPRaF-N \cite{zhang2014oblique}, RaF-PCA \cite{zhang2014random}, RaF-LDA \cite{zhang2014random}, DRaF \cite{han2020double}, MPDRaF-T, MPDRaF-P, MPDRaF-N, DRaF-PCA and DRaF-LDA classification models } \label{table:Bias_Varaince_Table} \\
\hline
Datasets &RaF &MPRaF-T&MPRaF-P&MPRaF-N&RaF-PCA&RaF-LDA&DRaF&MPDRaF-T$^*$&MPDRaF-P$^*$&MPDRaF-N$^*$&DRaF-PCA$^*$&DRaF-LDA$^*$\\
 &Bias &Bias&Bias&Bias&Bias&Bias&Bias&Bias&Bias&Bias&Bias&Bias\\
&Variance &Variance&Variance&Variance&Variance&Variance&Variance&Variance&Variance&Variance&Variance&Variance\\
\hline 
\endfirsthead
\multicolumn{13}{c}%
{{\bfseries \tablename\ \thetable{} -- continued from previous page}} \\
\hline
Datasets &RaF &MPRaF-T&MPRaF-P&MPRaF-N&RaF-PCA&RaF-LDA&DRaF&MPDRaF-T$^*$&MPDRaF-P$^*$&MPDRaF-N$^*$&DRaF-PCA$^*$&DRaF-LDA$^*$\\
&Bias &Bias&Bias&Bias&Bias&Bias&Bias&Bias&Bias&Bias&Bias&Bias\\
&Variance &Variance&Variance&Variance&Variance&Variance&Variance&Variance&Variance&Variance&Variance&Variance\\
 \hline 
\endhead
\hline \multicolumn{13}{|r|}{{Continued on next page}} \\ \hline
\endfoot
\endlastfoot

\hline
       abalone&$422.19$&$421.95$&$422.16$&$409.26$&$424.42$&$420.62$&$415.77$&$411.11$&$417.33$&$399.97$&$415.42$&$413.34$\\&$220.38$&$224.11$&$226.61$&$206.18$&$228.12$&$225.96$&$202.89$&$204.21$&$214.17$&$190.48$&$215.17$&$214.64$\\
acute-inflammation&$2.21$&$1.48$&$2.14$&$1.99$&$1.72$&$1.74$&$1.81$&$1.44$&$1.64$&$2.07$&$1.56$&$1.53$\\&$1.84$&$1.31$&$1.84$&$1.74$&$1.52$&$1.48$&$1.53$&$1.25$&$1.4$&$1.83$&$1.31$&$1.32$\\
acute-nephritis&$1.14$&$0.86$&$1.14$&$1.53$&$0.74$&$0.72$&$0.53$&$0.86$&$0.71$&$1.18$&$0.58$&$0.53$\\&$0.99$&$0.8$&$1.02$&$1.41$&$0.66$&$0.66$&$0.49$&$0.8$&$0.66$&$1.09$&$0.54$&$0.5$\\
adult&$3032.87$&$3261.33$&$3192.13$&$3669.3$&$3341.57$&$3802.34$&$2798.83$&$2993.5$&$2967.71$&$3432.11$&$3175.48$&$3539.3$\\&$1423.89$&$1552.82$&$1555.3$&$1908.44$&$1718.95$&$2123.73$&$1110.31$&$1241.92$&$1287.36$&$1646.2$&$1553.7$&$1886.05$\\
annealing&$56.94$&$76.64$&$54.6$&$80.44$&$62.27$&$63.7$&$52.45$&$78.1$&$58.66$&$80.84$&$63.66$&$64.95$\\&$24.71$&$23.7$&$25.63$&$21.81$&$33.16$&$34.4$&$23.87$&$21.74$&$24.56$&$22.91$&$32.91$&$34.78$\\
arrhythmia&$48.16$&$55.61$&$47.75$&$54.86$&$53.39$&$53.65$&$45.18$&$52.32$&$45.5$&$50.95$&$49.35$&$50.5$\\&$30.55$&$35.78$&$30.37$&$35.2$&$34.99$&$35.65$&$26.16$&$30.44$&$26.79$&$28.58$&$29.82$&$32.12$\\
audiology-std&$11.27$&$14.81$&$10.96$&$15.59$&$15.04$&$14.59$&$11$&$14.25$&$11.12$&$15.71$&$14.47$&$14.25$\\&$6.9$&$10.19$&$6.73$&$10.3$&$10.22$&$9.98$&$5.79$&$9.92$&$6$&$10.03$&$9.82$&$9.83$\\
balance-scale&$34.84$&$30.63$&$31.25$&$30.69$&$32.12$&$30.29$&$32.76$&$28.41$&$28.69$&$27.46$&$30.59$&$27.93$\\&$22.44$&$20.42$&$20.46$&$20.4$&$21.08$&$19.99$&$17.75$&$17.64$&$17.18$&$16.99$&$17.8$&$16.84$\\
balloons&$1.42$&$1.42$&$1.41$&$1.35$&$1.43$&$1.33$&$1.26$&$1.19$&$1.27$&$1.31$&$1.11$&$1.14$\\&$0.8$&$0.82$&$0.82$&$0.83$&$0.84$&$0.77$&$0.66$&$0.67$&$0.65$&$0.75$&$0.65$&$0.66$\\
bank&$158.45$&$171.54$&$164.99$&$165.29$&$168.34$&$166.13$&$146.42$&$151.74$&$152.84$&$148.85$&$155.18$&$153.4$\\&$79.65$&$85.26$&$83.55$&$76.1$&$87.21$&$84.7$&$67.42$&$64.8$&$70.35$&$59.45$&$75$&$73.58$\\
blood&$52.18$&$50.54$&$51.52$&$50.64$&$52.67$&$52.28$&$47.96$&$44.88$&$48.01$&$46.45$&$49.84$&$49.51$\\&$19.29$&$18.11$&$19.18$&$19.06$&$19.8$&$19.75$&$11.12$&$10.79$&$11.85$&$12.1$&$13.16$&$12.94$\\
breast-cancer&$24.34$&$24.42$&$24.36$&$23.73$&$24.75$&$24.51$&$21.58$&$22.02$&$21.87$&$20.98$&$22.56$&$22.05$\\&$11.77$&$11.85$&$11.78$&$11.64$&$12.35$&$12.23$&$8.8$&$9.28$&$9$&$8.51$&$9.87$&$9.6$\\
breast-cancer-wisc&$9.94$&$9.37$&$8.91$&$9.06$&$9.99$&$9.29$&$9.31$&$8.12$&$8.07$&$8.01$&$8.91$&$8.75$\\&$5.92$&$5.6$&$5.22$&$4.98$&$6.03$&$5.35$&$5.25$&$4.5$&$4.5$&$4.38$&$4.99$&$4.89$\\
breast-cancer-wisc-diag&$11.58$&$9.6$&$10.32$&$10.57$&$13.13$&$10.09$&$10.13$&$8.5$&$8.84$&$8.47$&$11.23$&$8.46$\\&$6.82$&$5.79$&$6.28$&$6.57$&$8.5$&$6.47$&$5.65$&$4.93$&$5.16$&$4.91$&$7.17$&$5.07$\\
breast-cancer-wisc-prog&$16.76$&$15.82$&$16.42$&$16.32$&$17.21$&$16.64$&$15.98$&$14.64$&$15.08$&$14.58$&$15.74$&$16.16$\\&$8.9$&$8.65$&$8.63$&$8.73$&$9.24$&$8.94$&$8.11$&$7.55$&$7.98$&$7.41$&$8.29$&$8.43$\\
breast-tissue&$9.3$&$9.45$&$9.77$&$9.95$&$10.14$&$9.38$&$8.33$&$9.1$&$8.89$&$9.2$&$8.96$&$8.71$\\&$5.28$&$5.81$&$5.84$&$6.32$&$6.47$&$5.91$&$4.45$&$4.87$&$4.85$&$5.37$&$5.18$&$4.9$\\
car&$53.62$&$66.26$&$57.3$&$82.54$&$57.16$&$51.75$&$47.24$&$60.48$&$48.71$&$76.95$&$46.33$&$42.38$\\&$40.16$&$49.16$&$43.61$&$52.95$&$43.64$&$39.4$&$35.52$&$44.72$&$37.55$&$46.94$&$35.5$&$33.03$\\
cardiotocography-10clases&$131.33$&$173.39$&$137.05$&$184.31$&$167.74$&$146.51$&$109.53$&$152.42$&$116.46$&$171.16$&$142.7$&$125.47$\\&$89.8$&$125.16$&$96.6$&$132.33$&$124.55$&$106.39$&$70.11$&$106.29$&$76.81$&$121.06$&$102.89$&$86.74$\\
cardiotocography-3clases&$57.33$&$73.66$&$62.12$&$78.34$&$73.06$&$69.02$&$47.51$&$62.48$&$51.28$&$68.54$&$62.08$&$57.57$\\&$37.12$&$47.85$&$40.8$&$50.63$&$47.74$&$45.32$&$28.08$&$38.44$&$31.89$&$42.3$&$38.38$&$36.22$\\
chess-krvk&$3034.05$&$3362.72$&$3124.59$&$3727.55$&$2940.16$&$2994.79$&$2827.31$&$3183.27$&$2887.18$&$3648.43$&$2662.02$&$2724.99$\\&$2059.91$&$2326.91$&$2178.24$&$2416.27$&$2080.07$&$2093.71$&$1831.25$&$2132.02$&$1957.51$&$2302.68$&$1831.15$&$1848.16$\\
chess-krvkp&$80.61$&$120.16$&$83.73$&$155.37$&$102.15$&$90.43$&$58.94$&$98.25$&$64.73$&$138.61$&$79.92$&$73.46$\\&$61.57$&$90.21$&$64.48$&$113.06$&$77.99$&$69.75$&$44.63$&$74.42$&$50.15$&$102.97$&$61.82$&$57.61$\\
congressional-voting&$44.37$&$44.87$&$45.13$&$44.68$&$46.33$&$45.13$&$42.7$&$42.79$&$42.63$&$43.49$&$43.21$&$42.87$\\&$16.03$&$14.65$&$15.34$&$13.96$&$18.56$&$17.31$&$3.25$&$3.8$&$3.4$&$6.23$&$5.88$&$4.65$\\
conn-bench-sonar-mines-rocks&$17.83$&$17.62$&$18.6$&$18$&$19.54$&$17.24$&$16.33$&$16.47$&$16.78$&$16.82$&$17.18$&$15.5$\\&$10.12$&$10.15$&$10.55$&$10.38$&$11.02$&$9.62$&$9.37$&$9.19$&$9.77$&$9.4$&$9.89$&$8.92$\\
conn-bench-vowel-deterding&$116.57$&$95.58$&$112.38$&$124.23$&$121.65$&$109.1$&$76.15$&$60.54$&$70.67$&$101.47$&$78.76$&$68.95$\\&$105.4$&$92.13$&$105.05$&$117.06$&$113.98$&$102.3$&$71.13$&$61.26$&$69.17$&$99.09$&$78.68$&$68.13$\\
connect-4&$3786.62$&$4350.55$&$3949.26$&$4354.23$&$4067.8$&$3991.2$&$3372.93$&$4027.03$&$3573.11$&$4160.29$&$3628$&$3581.14$\\&$1936.33$&$1787.5$&$1932.43$&$1384.57$&$2150.1$&$2055.04$&$1546.41$&$1353.85$&$1556.88$&$1148.35$&$1772.59$&$1673.69$\\
contrac&$178$&$186.21$&$182.14$&$187$&$181.85$&$181.28$&$175.47$&$182.93$&$178.55$&$180.44$&$179.98$&$179.21$\\&$91.6$&$96.67$&$94.68$&$98.58$&$94.61$&$93.99$&$75.21$&$82$&$80.1$&$87.81$&$80.62$&$80.66$\\
credit-approval&$35.78$&$38.42$&$38.92$&$37.32$&$39.4$&$38.34$&$33.89$&$34.78$&$34.9$&$35.15$&$37.12$&$35.42$\\&$19.87$&$21.82$&$22.12$&$20.93$&$22.69$&$21.69$&$17.87$&$17.97$&$19.05$&$18.67$&$20.57$&$19.26$\\
cylinder-bands&$44.55$&$49.18$&$46.05$&$48.93$&$47.79$&$46.45$&$40.46$&$47.67$&$42.35$&$46.89$&$44.39$&$43.68$\\&$25.53$&$27.29$&$26.23$&$26.87$&$27.08$&$26.09$&$23.47$&$26.66$&$24.57$&$26.55$&$25.48$&$24.91$\\
dermatology&$9.98$&$10.83$&$9.38$&$11.47$&$13.46$&$11.9$&$7.52$&$9.16$&$7.64$&$9.71$&$10.38$&$9.35$\\&$8.43$&$9.44$&$7.97$&$9.97$&$11.84$&$10.41$&$5.96$&$7.71$&$6.03$&$8.33$&$9.03$&$8.01$\\
echocardiogram&$9.15$&$8.86$&$9.1$&$9.38$&$9.44$&$9.19$&$8.03$&$7.87$&$8.28$&$8.04$&$8.78$&$8.11$\\&$4.68$&$4.69$&$4.87$&$4.96$&$5.04$&$4.95$&$3.72$&$3.77$&$4.15$&$3.79$&$4.38$&$4.03$\\
ecoli&$18.23$&$19.64$&$18.28$&$20.9$&$19.4$&$18.85$&$15.41$&$17.79$&$15.86$&$19.05$&$16.9$&$16.7$\\&$10.84$&$11.67$&$10.89$&$13.62$&$12.08$&$11.52$&$7.47$&$9.45$&$8.21$&$11.62$&$8.97$&$8.87$\\
energy-y1&$16.04$&$24.28$&$17.47$&$23.22$&$20.83$&$17.95$&$12.73$&$19.82$&$13.21$&$20.27$&$15.18$&$13.49$\\&$8.06$&$14.04$&$9.48$&$13.82$&$12.6$&$10.71$&$5.35$&$9.79$&$5.86$&$10.98$&$8.25$&$7.04$\\
energy-y2&$23.08$&$26.22$&$23.5$&$27.07$&$24.83$&$23.9$&$22.61$&$23.18$&$21.97$&$26.04$&$23.07$&$22.62$\\&$10.96$&$13.58$&$11.08$&$14.46$&$12.98$&$12.13$&$7.2$&$9.18$&$7.58$&$12.87$&$9.78$&$8.9$\\
fertility&$4.31$&$4.93$&$4.55$&$4.56$&$4.73$&$4.57$&$3.67$&$3.86$&$3.91$&$3.7$&$3.9$&$3.95$\\&$1.94$&$2.47$&$2.06$&$2.06$&$2.27$&$2.09$&$1.12$&$1.31$&$1.38$&$1.08$&$1.43$&$1.47$\\
flags&$21.28$&$25.33$&$22.76$&$25.36$&$24.8$&$25$&$19.63$&$24.64$&$20.77$&$24.74$&$23.38$&$23.12$\\&$14.72$&$16.74$&$15.71$&$17.01$&$16.75$&$17.04$&$12.68$&$15.6$&$13.78$&$15.65$&$15.34$&$15.29$\\
glass&$19.46$&$20.25$&$19.09$&$20.78$&$20.5$&$20.51$&$16.67$&$18.63$&$17.23$&$19.99$&$18.63$&$18.57$\\&$12.45$&$12.55$&$12.2$&$12.87$&$13.1$&$12.98$&$9.7$&$10.73$&$10.14$&$11.45$&$11.37$&$11.18$\\
haberman-survival&$25.47$&$25.79$&$25.48$&$25.05$&$26.5$&$26.28$&$25.04$&$24.18$&$24.92$&$24.09$&$25.51$&$25.24$\\&$11.01$&$11.38$&$11$&$10.92$&$11.64$&$11.61$&$7.83$&$8.31$&$8.69$&$8.14$&$8.86$&$9.16$\\
hayes-roth&$5.92$&$9.26$&$8.03$&$10.3$&$8.09$&$8.58$&$4.57$&$7.82$&$5.92$&$9.55$&$6.86$&$7.24$\\&$3.43$&$5.99$&$5.21$&$6.48$&$5.45$&$5.44$&$2.37$&$4.98$&$3.56$&$5.76$&$4.52$&$4.61$\\
heart-cleveland&$32.34$&$32.18$&$31.56$&$32.01$&$32.1$&$31.78$&$31.08$&$30.46$&$30.52$&$30.2$&$31$&$31.05$\\&$19.01$&$19.59$&$18.77$&$19.27$&$19.53$&$19.35$&$16.28$&$16.73$&$16.64$&$16.38$&$17.69$&$17.68$\\
heart-hungarian&$17.72$&$17.95$&$18.06$&$17.68$&$18.68$&$17.26$&$15.66$&$15.94$&$16.59$&$15.6$&$16.65$&$16.44$\\&$9.05$&$9.54$&$9.56$&$9.41$&$10.05$&$9.19$&$7.06$&$7.67$&$7.92$&$7.52$&$8.34$&$8.12$\\
heart-switzerland&$17.46$&$17.32$&$17.33$&$17.45$&$17.62$&$17.52$&$17.16$&$17.5$&$17.15$&$17.44$&$17.58$&$17.24$\\&$9.94$&$10.18$&$10.02$&$10.2$&$10.31$&$10.23$&$9.04$&$9.31$&$8.9$&$9.19$&$9.78$&$9.66$\\
heart-va&$29.78$&$29.76$&$29.87$&$29.57$&$29.55$&$29.72$&$29.61$&$30.17$&$29.46$&$29.33$&$29.59$&$29.76$\\&$17.39$&$17.41$&$17.36$&$17.5$&$17.44$&$17.49$&$15.93$&$16.11$&$16.02$&$16.16$&$16.35$&$16.34$\\
hepatitis&$8.91$&$9.79$&$9.41$&$9.57$&$9.48$&$9.05$&$8.82$&$9.19$&$8.89$&$9.06$&$9.04$&$8.96$\\&$4.91$&$5.54$&$5.26$&$5.5$&$5.45$&$5.18$&$4.71$&$5.11$&$4.81$&$5.04$&$4.92$&$4.88$\\
hill-valley&$293.58$&$272.38$&$281$&$268.79$&$281.38$&$273.96$&$290.39$&$264.8$&$274.21$&$261.01$&$271.99$&$267.73$\\&$135.21$&$143.5$&$144.06$&$137.19$&$145$&$138.52$&$132.08$&$141.33$&$142.91$&$131.17$&$142.78$&$137.4$\\
horse-colic&$18.23$&$20.99$&$19.67$&$20.73$&$22.68$&$22.33$&$16.51$&$19.16$&$18.7$&$20.2$&$21.22$&$22.24$\\&$10.34$&$11.9$&$11.27$&$11.99$&$12.96$&$12.69$&$8.78$&$10.51$&$10.31$&$11.22$&$11.82$&$12.32$\\
ilpd-indian-liver&$48.37$&$48.88$&$49.08$&$50.59$&$48.99$&$49.22$&$46.68$&$46.75$&$46.17$&$47.11$&$46.68$&$47.47$\\&$23.29$&$23.55$&$23.93$&$24.73$&$24.41$&$24.09$&$20.95$&$20.99$&$21.37$&$21.25$&$22.14$&$22.45$\\
image-segmentation&$380.59$&$466.36$&$424.88$&$623.46$&$457.6$&$468.05$&$298.42$&$393.75$&$338.4$&$588.04$&$378.66$&$392.93$\\&$295.25$&$377.07$&$348.3$&$520.25$&$375.6$&$398.97$&$227.28$&$315.87$&$270.57$&$490.47$&$315.18$&$339.2$\\
ionosphere&$13.06$&$14.02$&$14.4$&$14.62$&$14.84$&$13.84$&$10.89$&$12.21$&$12.45$&$12.83$&$13.34$&$12.7$\\&$7.97$&$9.19$&$9.42$&$9.32$&$9.71$&$8.95$&$6.4$&$8$&$8.19$&$8.4$&$8.69$&$8$\\
iris&$2.02$&$2.11$&$1.77$&$2.02$&$2.77$&$1.55$&$1.95$&$1.52$&$1.59$&$1.39$&$2.28$&$1.54$\\&$1.08$&$1.33$&$1.13$&$1.42$&$2.02$&$0.88$&$0.84$&$0.78$&$0.86$&$0.81$&$1.46$&$0.88$\\
led-display&$79.9$&$84.24$&$80.89$&$90.26$&$78.53$&$79.14$&$78.8$&$83.67$&$80.07$&$90.53$&$77.87$&$78.75$\\&$38.74$&$45.57$&$40.56$&$55.39$&$38.18$&$38.21$&$28.95$&$35.1$&$31.03$&$49.77$&$26.71$&$29.18$\\
lenses&$1.93$&$1.85$&$1.77$&$2.06$&$2.13$&$1.96$&$1.64$&$1.8$&$1.94$&$1.57$&$1.9$&$1.79$\\&$1.32$&$1.28$&$1.17$&$1.31$&$1.39$&$1.33$&$1.02$&$1.15$&$1.17$&$1.05$&$1.2$&$1.14$\\
letter&$875.1$&$1169.82$&$1122.19$&$1298.45$&$1182.43$&$1040.52$&$684.66$&$988.71$&$890.24$&$1159.74$&$940.64$&$816.4$\\&$764.88$&$1047.64$&$1001.99$&$1159.21$&$1055.78$&$932.48$&$594$&$890.77$&$802.66$&$1044.75$&$852.52$&$743.86$\\
libras&$39.03$&$34$&$39.06$&$36.55$&$38.12$&$37.04$&$30.97$&$27.25$&$31.19$&$30.93$&$29.77$&$28.47$\\&$29.95$&$27.67$&$31.79$&$29.51$&$30.76$&$29.78$&$23.66$&$22.48$&$25.89$&$25.5$&$24.2$&$23.26$\\
low-res-spect&$21.91$&$21.87$&$23.23$&$26.01$&$25.24$&$23.35$&$20.2$&$20.31$&$21.19$&$23.63$&$21.42$&$19.92$\\&$14.56$&$14.63$&$16.17$&$18.25$&$18.37$&$16.53$&$12.33$&$12.7$&$13.87$&$16.04$&$14.45$&$13.34$\\
lung-cancer&$4.26$&$4.17$&$4.33$&$4.34$&$4.44$&$4.39$&$3.98$&$4.31$&$4.06$&$4.23$&$4.2$&$4.26$\\&$2.41$&$2.45$&$2.38$&$2.48$&$2.55$&$2.51$&$2.14$&$2.45$&$2.13$&$2.43$&$2.43$&$2.39$\\
lymphography&$29.68$&$29.16$&$29.39$&$28.59$&$29.28$&$30$&$30.62$&$29.47$&$30.51$&$29.1$&$29.72$&$29.84$\\&$6.14$&$7.17$&$6.35$&$7.45$&$7.16$&$6.57$&$4.95$&$6.21$&$5.02$&$6.48$&$6.17$&$5.89$\\
magic&$966.3$&$1008.49$&$1008.6$&$1004.6$&$996.78$&$976.48$&$898$&$916.77$&$941.86$&$896.96$&$920.04$&$910.56$\\&$505.97$&$533.54$&$537.45$&$534.69$&$542.88$&$522.07$&$448$&$448.36$&$481.86$&$437.62$&$486.41$&$474.26$\\
mammographic&$54.54$&$53.46$&$57.6$&$52.97$&$56.38$&$57.43$&$52.55$&$50.14$&$52.68$&$50.64$&$52.91$&$53.6$\\&$23.19$&$21.7$&$24.86$&$21.59$&$23.8$&$24.96$&$14.66$&$15.31$&$16.96$&$17.48$&$16.02$&$16.37$\\
miniboone&$4121.21$&$3859.11$&$4216.63$&$3803.82$&$4551.71$&$4175.43$&$3858.53$&$3504.21$&$3965.63$&$3479.97$&$4206.87$&$3913.31$\\&$2422.28$&$2162.95$&$2504.94$&$2050.18$&$2765.54$&$2493.98$&$2254.23$&$1914.7$&$2351.13$&$1853.21$&$2542.82$&$2335.05$\\
molec-biol-promoter&$9.42$&$10.33$&$9.78$&$10.49$&$11.02$&$9.94$&$8.33$&$9.79$&$9.17$&$9.58$&$10.18$&$9.34$\\&$5.6$&$5.9$&$5.76$&$6.01$&$6.1$&$5.79$&$5.11$&$5.66$&$5.46$&$5.6$&$5.87$&$5.53$\\
molec-biol-splice&$181.66$&$290.79$&$213.17$&$297.37$&$306.32$&$244.77$&$145.8$&$261.31$&$180.77$&$267.25$&$269.7$&$214.22$\\&$138.02$&$201.95$&$159.45$&$205.03$&$212.57$&$177.08$&$109.67$&$182.44$&$135.65$&$187.34$&$189.93$&$154.95$\\
monks-1&$184.7$&$188.87$&$190.15$&$193.74$&$190.17$&$189.89$&$179.06$&$181.61$&$184.26$&$191.03$&$188.55$&$182.31$\\&$52.93$&$87.64$&$69.57$&$80.62$&$77.94$&$75.1$&$47.2$&$75.36$&$67.2$&$83.85$&$68.68$&$66.87$\\
monks-2&$153.7$&$157.68$&$155.9$&$162.69$&$163.26$&$160.78$&$148.77$&$150.26$&$151.06$&$154.24$&$154.44$&$154.19$\\&$30.91$&$44.29$&$38.65$&$54.25$&$51.21$&$47.28$&$16.71$&$26.01$&$23.34$&$35.95$&$33.11$&$30.23$\\
monks-3&$194.02$&$172.46$&$188.45$&$184.22$&$188.32$&$192.86$&$200.1$&$169.54$&$186.25$&$188.42$&$181.59$&$190.13$\\&$57.03$&$73.83$&$61.84$&$74.99$&$76.46$&$74.71$&$41.86$&$66.7$&$57.22$&$63.91$&$59.58$&$66.64$\\
mushroom&$3.18$&$6.41$&$3.36$&$27.9$&$9.02$&$7.78$&$2.21$&$4.61$&$2.23$&$22.76$&$9.97$&$4.58$\\&$2.96$&$6.01$&$3.19$&$26.48$&$8.51$&$7.39$&$2.11$&$4.41$&$2.16$&$21.69$&$9.63$&$4.32$\\
musk-1&$36.02$&$37.26$&$39.74$&$38.63$&$41.52$&$38.78$&$30.47$&$32.76$&$34.15$&$34.71$&$35.13$&$34.42$\\&$21.41$&$22.39$&$23.37$&$23.26$&$24.97$&$22.56$&$17.96$&$19.98$&$20.56$&$21.08$&$21.42$&$20.02$\\
musk-2&$115.57$&$146.03$&$149.62$&$159.35$&$175.74$&$157.47$&$74.74$&$114.79$&$115.99$&$122.98$&$125.67$&$120.31$\\&$74.39$&$94.63$&$96.47$&$104.31$&$117.35$&$103.39$&$48.23$&$70.91$&$72.46$&$75.96$&$82.9$&$78.09$\\
nursery&$2030.12$&$1962.02$&$2014.15$&$1920.23$&$2004.34$&$2021.52$&$2058.12$&$1985.2$&$2046.31$&$1939.8$&$2050.82$&$2058$\\&$161.98$&$238.37$&$178.16$&$324.16$&$198.46$&$171.28$&$132.12$&$209.75$&$140.91$&$314.17$&$150.08$&$132.66$\\
OM\_nucleus\_4d&$81.66$&$74.48$&$76.05$&$72.5$&$78.9$&$74.33$&$76.11$&$68.82$&$71.97$&$65.53$&$74.18$&$68$\\&$42.31$&$40.62$&$41.99$&$37.79$&$44.34$&$42.02$&$38.87$&$37.36$&$39.42$&$33.5$&$41.79$&$37.88$\\
OM\_states\_2f&$34.01$&$34.28$&$34.58$&$35.49$&$37.72$&$32.94$&$31.06$&$30.45$&$31.41$&$31.39$&$33.8$&$29.99$\\&$19.83$&$19.9$&$20.75$&$21.12$&$23.55$&$20.02$&$17.22$&$17.43$&$18.2$&$18.1$&$20.25$&$17.49$\\
OT\_nucleus\_2f&$74.58$&$72.35$&$73.19$&$71.7$&$73.64$&$69.96$&$68.6$&$66.13$&$68.06$&$66.72$&$68.15$&$64.49$\\&$40.36$&$40.87$&$41.87$&$40.8$&$41.99$&$39.73$&$36.25$&$37.3$&$38.51$&$37.36$&$39.01$&$35.9$\\
OT\_states\_5b&$39.31$&$37.43$&$39.03$&$40.35$&$42.03$&$37.07$&$34.83$&$31.38$&$34.19$&$33.6$&$35.96$&$31.24$\\&$23.74$&$23.71$&$24.49$&$25.67$&$27.71$&$24.22$&$20.25$&$19.51$&$21.27$&$20.76$&$23.37$&$20.16$\\
optical&$347.14$&$471.68$&$351.36$&$932.06$&$525.93$&$854.82$&$293.25$&$424.12$&$301.73$&$941.64$&$473.15$&$746.27$\\&$311.63$&$429.48$&$317.48$&$724.69$&$482.75$&$698.24$&$265.14$&$391.18$&$272.43$&$729.29$&$443.37$&$643.02$\\
ozone&$30.14$&$30.46$&$30.64$&$29.86$&$35$&$33.74$&$28.36$&$26.49$&$27.92$&$24.97$&$30.81$&$30.67$\\&$15$&$16.01$&$15.45$&$15.3$&$19.46$&$18.5$&$13.23$&$11.87$&$13.25$&$10.05$&$15.93$&$15.73$\\
page-blocks&$61.43$&$65.49$&$62.18$&$71.87$&$67.36$&$65.3$&$54.61$&$56.4$&$55.72$&$65.77$&$58.29$&$56.73$\\&$34.01$&$36.47$&$35.83$&$42.22$&$39.94$&$38.53$&$24.99$&$27.94$&$27.59$&$36.35$&$30.5$&$29.61$\\
parkinsons&$9.43$&$9.15$&$9.53$&$9.47$&$10.49$&$9.77$&$7.89$&$7.72$&$8.06$&$8.14$&$8.98$&$8.24$\\&$5.16$&$5.53$&$5.5$&$5.56$&$6.1$&$6.06$&$4.3$&$4.73$&$4.78$&$4.91$&$5.11$&$4.88$\\
pendigits&$432.84$&$349.16$&$411.69$&$461.57$&$442.92$&$392.22$&$371.49$&$296.03$&$353.34$&$415.33$&$390.87$&$340.92$\\&$334.57$&$280.17$&$325.47$&$379.57$&$375$&$322.83$&$281.79$&$233.79$&$278.04$&$338.84$&$328.11$&$276.92$\\
pima&$62.07$&$63.41$&$63.57$&$64.1$&$64.12$&$63$&$60.56$&$60.16$&$61.79$&$59.96$&$62.36$&$61.22$\\&$30.1$&$31.52$&$31.52$&$31.97$&$31.71$&$31.04$&$27.11$&$27.75$&$29.4$&$28.04$&$29.64$&$28.43$\\
pittsburg-bridges-MATERIAL&$4.32$&$3.93$&$4.51$&$4$&$4.56$&$4.33$&$3.65$&$3.28$&$3.64$&$3.27$&$3.69$&$3.34$\\&$2.45$&$2.53$&$2.76$&$2.51$&$2.96$&$2.71$&$1.56$&$1.71$&$1.58$&$1.7$&$2.03$&$1.62$\\
pittsburg-bridges-REL-L&$9.87$&$9.77$&$9.89$&$9.77$&$10.05$&$10.19$&$9.04$&$8.99$&$9.16$&$8.98$&$9.28$&$9.05$\\&$5.66$&$5.84$&$5.88$&$5.88$&$5.92$&$6.01$&$4.78$&$4.92$&$4.75$&$5$&$5.09$&$4.96$\\
pittsburg-bridges-SPAN&$9.73$&$9.39$&$9.7$&$9.57$&$9.61$&$9.55$&$9.36$&$8.72$&$9.08$&$8.87$&$9.26$&$9.19$\\&$5.02$&$4.94$&$5$&$5.26$&$5.25$&$5.11$&$3.85$&$3.88$&$3.9$&$3.98$&$4.27$&$4.19$\\
pittsburg-bridges-T-OR-D&$3.8$&$4.23$&$3.76$&$3.95$&$4.11$&$4.08$&$3.13$&$3.13$&$3.18$&$3.29$&$3.6$&$3.51$\\&$1.89$&$2.16$&$1.87$&$2$&$2.12$&$2.03$&$1.18$&$1.05$&$1.19$&$1.24$&$1.51$&$1.45$\\
pittsburg-bridges-TYPE&$11.06$&$11.16$&$11.43$&$11.38$&$11.39$&$11.44$&$10.23$&$9.92$&$10.1$&$10.34$&$10.45$&$10.18$\\&$6.52$&$7.2$&$6.83$&$7.13$&$7.19$&$7.09$&$4.72$&$5.46$&$5.01$&$5.61$&$5.67$&$5.47$\\
planning&$18.94$&$18.72$&$19.12$&$18.83$&$19.18$&$18.78$&$18.43$&$17.87$&$17.96$&$16.92$&$18.36$&$18.3$\\&$8.34$&$8.66$&$8.61$&$8.58$&$8.95$&$8.67$&$7.44$&$7.37$&$7.76$&$6.7$&$7.82$&$7.91$\\
plant-margin&$207.32$&$213.59$&$225.57$&$224.65$&$214.8$&$220.86$&$185.65$&$203.73$&$208.48$&$217.32$&$189.21$&$191.76$\\&$165.7$&$168.02$&$175.83$&$174.65$&$169.32$&$173.42$&$150.92$&$162.44$&$166.6$&$170.34$&$153.6$&$156.81$\\
plant-shape&$208.39$&$200.15$&$218.05$&$213.26$&$207.39$&$210.35$&$187.89$&$186.74$&$199.96$&$201.49$&$181.49$&$179.02$\\&$151.83$&$151.54$&$165.52$&$162.81$&$157.17$&$161.98$&$131.84$&$142.14$&$152.98$&$154.54$&$136.95$&$138.34$\\
plant-texture&$200.01$&$204.59$&$217.09$&$218.8$&$209.02$&$215.93$&$172.62$&$191.51$&$197.74$&$208.96$&$183.31$&$187.08$\\&$161.71$&$165.22$&$173.28$&$173.36$&$167.88$&$172.22$&$142.61$&$156.98$&$161.7$&$167.84$&$150.72$&$155.16$\\
post-operative&$9.14$&$9.2$&$9.15$&$9.45$&$9.25$&$9.94$&$7.73$&$8.02$&$7.79$&$8.18$&$8.5$&$8.18$\\&$3.58$&$3.53$&$3.47$&$3.78$&$3.54$&$3.92$&$1.69$&$1.94$&$1.81$&$2.07$&$2.28$&$2.12$\\
primary-tumor&$41.25$&$43.21$&$41.32$&$43.23$&$42.36$&$41.73$&$38.62$&$40.04$&$38.99$&$40.65$&$39.66$&$39.88$\\&$26.24$&$28.4$&$26.7$&$29.11$&$27.75$&$27.34$&$20.98$&$23.19$&$21.99$&$25.14$&$23.35$&$24.13$\\
ringnorm&$256.67$&$306.79$&$304.97$&$304.71$&$263.7$&$276.46$&$213.73$&$285.23$&$281$&$275.01$&$220.26$&$231.47$\\&$175.88$&$168.89$&$170.48$&$169.11$&$191.62$&$198.7$&$141.68$&$163.36$&$161.3$&$159.62$&$160.21$&$167$\\
seeds&$6.82$&$5.48$&$6.61$&$6.06$&$7.12$&$5.76$&$5.84$&$5.11$&$5.38$&$5.35$&$6.1$&$5.2$\\&$3.77$&$3.24$&$3.87$&$3.42$&$4.32$&$3.24$&$3.1$&$2.75$&$3.02$&$2.81$&$3.47$&$2.92$\\
semeion&$126.93$&$163.75$&$139.3$&$167.68$&$162.94$&$144.83$&$106.49$&$150.44$&$118.57$&$154.24$&$138.95$&$121.39$\\&$108.94$&$135.69$&$118.51$&$138.09$&$134.91$&$123.66$&$91.81$&$126.4$&$102.45$&$129.21$&$118.36$&$106.29$\\
soybean&$85.7$&$125.43$&$86.86$&$169.31$&$145.14$&$168.43$&$65.71$&$118.51$&$66.27$&$167.24$&$129.05$&$160.1$\\&$70.2$&$109.22$&$71.29$&$140.08$&$126.74$&$141.95$&$49.24$&$104.9$&$50.77$&$139.17$&$115.31$&$137.44$\\
spambase&$135.46$&$161.62$&$143.8$&$179.6$&$156.4$&$149.42$&$113.87$&$140.49$&$121.61$&$156.76$&$131.21$&$128.29$\\&$83.45$&$104.31$&$90.36$&$117.42$&$101.55$&$95.58$&$66.39$&$87.83$&$73.1$&$99.73$&$82.8$&$80.36$\\
spect&$74.55$&$78.75$&$75.48$&$81.66$&$81.93$&$79.26$&$70.13$&$75.54$&$71.13$&$76.95$&$77.76$&$76.27$\\&$35.82$&$37.59$&$36.46$&$37.47$&$39.29$&$36.74$&$28.16$&$29.38$&$28.01$&$29.5$&$32.29$&$32.7$\\
spectf&$37.38$&$44.39$&$42.58$&$28.7$&$50.73$&$60.26$&$33.26$&$39.24$&$36.48$&$23$&$45.36$&$47.13$\\&$22.54$&$28.01$&$26.71$&$15.24$&$32.63$&$37.78$&$18.92$&$24.42$&$21.91$&$9.16$&$28.63$&$29.82$\\
statlog-australian-credit&$72.45$&$76.02$&$74.31$&$72.89$&$75.22$&$75.5$&$73.17$&$75.86$&$75.58$&$75.18$&$75.13$&$75.66$\\&$35.72$&$36.79$&$36.61$&$35.72$&$36.67$&$36.65$&$35.08$&$36.52$&$36.82$&$35.68$&$35.9$&$36.08$\\
statlog-german-credit&$82.59$&$86.24$&$83.99$&$85.12$&$85.53$&$83.99$&$76.37$&$80.24$&$79.5$&$80.47$&$80.94$&$79.28$\\&$43.05$&$44.83$&$43.79$&$43.73$&$44.69$&$43.93$&$37.2$&$38.65$&$39.35$&$38.63$&$40.84$&$39.73$\\
statlog-heart&$17.4$&$17.22$&$17.67$&$17.85$&$17.62$&$17$&$16.33$&$15.9$&$16.24$&$16.6$&$16.49$&$16.19$\\&$9.81$&$9.82$&$10$&$10.02$&$9.98$&$9.37$&$8.18$&$8.39$&$8.55$&$8.77$&$9.02$&$8.69$\\
statlog-image&$42.92$&$53.07$&$46.48$&$67.83$&$51.13$&$42.32$&$30.38$&$41.77$&$34.76$&$53.36$&$36.73$&$30.74$\\&$32.37$&$41.95$&$36.4$&$54.87$&$41.79$&$33.69$&$22.42$&$32.39$&$26.73$&$42.74$&$29.24$&$23.59$\\
statlog-landsat&$330.98$&$325.12$&$327.45$&$343.26$&$351.89$&$334.89$&$301.92$&$298.76$&$298.83$&$317.8$&$316.78$&$306.57$\\&$215.27$&$205.78$&$210.92$&$213.63$&$232.29$&$215.5$&$191.15$&$186.25$&$188.09$&$193.7$&$205.19$&$194.98$\\
statlog-shuttle&$14.63$&$62.49$&$26.19$&$169.51$&$59.99$&$77.66$&$8.09$&$69.82$&$21.17$&$159.44$&$55.01$&$53.09$\\&$10.99$&$51.74$&$21.32$&$145.12$&$54.77$&$70.31$&$6.06$&$58.72$&$16.2$&$134.38$&$51.13$&$48.18$\\
statlog-vehicle&$68.72$&$68.8$&$67.81$&$70.73$&$72.12$&$67.28$&$62.6$&$64.3$&$62.07$&$65.36$&$65.54$&$60.85$\\&$41.53$&$43.94$&$43.43$&$45.73$&$48.4$&$44.34$&$35.74$&$39.18$&$38.16$&$41.03$&$42.64$&$39.28$\\
steel-plates&$158.94$&$168.25$&$167.32$&$176.17$&$174.42$&$169.94$&$143.81$&$157.85$&$150.39$&$163.29$&$156.3$&$156.1$\\&$102.96$&$111.27$&$110.38$&$115.95$&$116.19$&$112.44$&$89.65$&$101.28$&$97.22$&$107.58$&$102.84$&$100.54$\\
synthetic-control&$24.33$&$20.63$&$24.89$&$28.59$&$27.24$&$21.6$&$18.64$&$16.33$&$19.11$&$22.57$&$20.27$&$17.68$\\&$19.91$&$17.23$&$20.25$&$23.2$&$21.5$&$17.6$&$15.27$&$13.42$&$15.76$&$18.25$&$16.36$&$14.55$\\
teaching&$18.3$&$18.48$&$18.38$&$18.72$&$18.77$&$18.5$&$17.13$&$17.95$&$17.34$&$18.54$&$17.15$&$17.22$\\&$9.98$&$10.03$&$10.09$&$10.11$&$10.23$&$9.88$&$7.15$&$7.57$&$7.47$&$8.44$&$7.54$&$7.61$\\
thyroid&$96.4$&$235.15$&$103.95$&$764.03$&$327.21$&$473.01$&$85.6$&$217.64$&$84.42$&$714.99$&$347.09$&$433.88$\\&$64.68$&$152.4$&$71.55$&$587.65$&$277.09$&$396.42$&$54.73$&$137.61$&$53.7$&$558.06$&$307.19$&$369.07$\\
tic-tac-toe&$47.05$&$49.97$&$49.88$&$53.77$&$51$&$50.03$&$37.48$&$39.81$&$40.57$&$47.01$&$39.98$&$39.69$\\&$33.66$&$35.23$&$35.05$&$36.53$&$35.73$&$35.22$&$27.83$&$29.77$&$30.07$&$33.41$&$29.71$&$29.61$\\
titanic&$119.26$&$120.25$&$119.76$&$122.17$&$118.52$&$118.64$&$118.31$&$119.18$&$118.84$&$121.15$&$118.62$&$117.76$\\&$13.4$&$15.83$&$14.35$&$17.7$&$12.64$&$12.65$&$7$&$11.63$&$8.12$&$16.16$&$8.8$&$7.18$\\
trains&$0.75$&$0.69$&$0.75$&$0.73$&$0.83$&$0.83$&$0.59$&$0.71$&$0.55$&$0.75$&$0.62$&$0.69$\\&$0.44$&$0.44$&$0.45$&$0.44$&$0.47$&$0.47$&$0.38$&$0.42$&$0.35$&$0.45$&$0.4$&$0.43$\\
twonorm&$322.13$&$174.14$&$179.14$&$175.13$&$196.7$&$174.59$&$300.72$&$154.18$&$163.55$&$152.77$&$173.41$&$156.68$\\&$228.03$&$123.14$&$128.34$&$123.87$&$140.69$&$123.59$&$212.1$&$107.54$&$114.87$&$106.02$&$122.73$&$109.67$\\
vertebral-column-2clases&$18.23$&$17.96$&$18.07$&$18.48$&$19.07$&$17.83$&$17.34$&$16.3$&$16.55$&$16.32$&$17.53$&$16.7$\\&$9.64$&$9.9$&$9.85$&$9.84$&$10.22$&$9.72$&$8.37$&$8.37$&$8.7$&$8.23$&$8.81$&$8.62$\\
vertebral-column-3clases&$18.24$&$19.38$&$18.97$&$19.38$&$19.65$&$18.88$&$16.69$&$17.36$&$17.24$&$17.18$&$18.4$&$16.97$\\&$10.39$&$11.58$&$11.49$&$11.48$&$11.94$&$11.35$&$9$&$9.59$&$9.7$&$9.74$&$10.76$&$9.47$\\
wall-following&$53.4$&$263.94$&$147.78$&$279.8$&$217.02$&$205.71$&$34.78$&$209.91$&$116.49$&$228.6$&$161.01$&$150.81$\\&$44.51$&$195.14$&$121.8$&$208.18$&$171.9$&$161.23$&$28.59$&$152.31$&$94.1$&$169.25$&$125.62$&$116.59$\\
waveform&$338.3$&$330.24$&$329.75$&$329.77$&$342.18$&$304.07$&$325.41$&$310.76$&$310.6$&$307.43$&$324.3$&$292.42$\\&$200.61$&$195.75$&$195.75$&$196.45$&$207.55$&$177.82$&$185.88$&$180.54$&$181.45$&$179.29$&$192.39$&$166.3$\\
waveform-noise&$365.29$&$396.83$&$373.3$&$402.08$&$402.48$&$346.01$&$346.85$&$370.86$&$356.66$&$371.62$&$373.77$&$323.59$\\&$229.31$&$250.02$&$234.66$&$254.27$&$259.66$&$215.75$&$212.51$&$231$&$220.16$&$232.58$&$237.8$&$199.23$\\
wine&$4.7$&$4.16$&$4.63$&$4.9$&$5.94$&$4.44$&$3.9$&$3.97$&$3.44$&$4.12$&$4.59$&$3.28$\\&$3.7$&$3.32$&$3.72$&$3.85$&$4.53$&$3.6$&$3.03$&$3.04$&$2.72$&$3.2$&$3.55$&$2.66$\\
wine-quality-red&$170.47$&$172.33$&$170.37$&$172.22$&$170.12$&$169.33$&$152.43$&$159.31$&$152.86$&$160.16$&$150.89$&$148.68$\\&$98.56$&$102.89$&$99.29$&$101.99$&$101.58$&$101.11$&$82.49$&$89.32$&$85.4$&$89.21$&$85.87$&$83.8$\\
wine-quality-white&$538.74$&$544.35$&$539.74$&$550.6$&$540.62$&$538.74$&$474.95$&$498.56$&$478.36$&$509.12$&$470.59$&$466.43$\\&$323.98$&$333.7$&$327.45$&$333.24$&$333.04$&$330.1$&$268.98$&$290.93$&$274.29$&$295.78$&$274.59$&$268.83$\\
yeast&$171.29$&$178.95$&$172.73$&$179.09$&$173.81$&$173.4$&$164.8$&$172.82$&$165.84$&$173.92$&$166.39$&$165.68$\\&$99.69$&$109.78$&$102.5$&$109.3$&$104.37$&$103.6$&$85.57$&$97.63$&$88.63$&$99.42$&$91.59$&$91.27$\\
zoo&$2.46$&$2.62$&$2.37$&$3.11$&$3.11$&$2.94$&$1.62$&$2.02$&$1.49$&$2.16$&$2.22$&$1.93$\\&$2.11$&$2.35$&$2.03$&$2.65$&$2.78$&$2.65$&$1.37$&$1.68$&$1.26$&$1.85$&$2.01$&$1.72$\\
\hline
    \multicolumn{13}{l}{Here, OM denotes oocytes\_merluccius, OT denotes oocytes\_trisopterus.}
    \end{longtable}
\end{landscape}
\restoregeometry

\begin{table}[]
    \centering
    \begin{tabular}{c|c|c|c}
    \hline
         Methods & Average Rank&Average Rank Difference&Significance \\
         \hline
         (RaF, DRaF)&$(7, 3.02)$&$3.98$ &Yes\\
         (MPRaF-T, MPDRaF-T)&$(8.52, 4.37)$&$4.15$& Yes\\
         (MPRaF-P, MPDRaF-P)&$(7.96, 3.55)$ &$4.41$& Yes\\
         (MPRaF-N, MPDRaF-N)&$(9.73, 5.9)$ &$3.83$& Yes\\
         (RaF-PCA, DRaF-PCA)&$(10.03, 5.5)$ &$4.53$& Yes\\
         (RaF-LDA, DRaF-LDA)&$(8.42, 4.01)$ &$4.41$& Yes\\
         \hline
    \end{tabular}\\
    $\chi^2_F=615.0719, F_F=103.0950,q_{0.05}=3.2680$. The two models are significantly different if the average ranks of the two models differ at least by the critical difference, $CD=1.5149$. 
    \caption{Significant difference among  the standard and double variants of the ensembles of decision trees based on the bias analysis.}
    \label{tab:bias_statitics}
\end{table}

\begin{table}[]
    \centering
    \begin{tabular}{c|c|c|c}
    \hline
          Methods & Average Rank&Average Rank Difference&Significance \\
         \hline
         (RaF, DRaF)&$(6.61, 1.93)$&$4.68$ &Yes\\
         (MPRaF-T, MPDRaF-T)&$(8.96, 4.03)$&$4.93$& Yes\\
         (MPRaF-P, MPDRaF-P)&$(8.13, 3.36)$ &$4.77$& Yes\\
         (MPRaF-N, MPDRaF-N)&$(9.88, 5.64)$ &$4.24$& Yes\\
         (RaF-PCA, DRaF-PCA)&$(10.68, 5.64)$ &$5.04$& Yes\\
         (RaF-LDA, DRaF-LDA)&$(8.93, 4.2)$ &$4.73$& Yes\\
         \hline
    \end{tabular}\\
    $\chi^2_F=809.8335, F_F=186.4664,q_{0.05}=3.2680$. The two models are significantly different if the average ranks of the two models differ at least by the critical difference, $CD=1.5149$. 
    \caption{Significant difference among  the standard and double variants of the ensembles of decision trees based on the variance analysis.}
    \label{tab:var_statistics}
\end{table}

\section{Conclusion}
\label{Sec:Conclusion}
In this paper, we propose two approaches for generating the double random forest models. In the first model,  we propose oblique double random forest ensemble models and in the second approach, we propose rotation based double random forest ensemble models.
 In oblique double random forest models, the splitting hyperplane at each non-leaf node is generated via MPSVM. This leads to the incorporation of geometric structure and hence, leads to better generalization performance. As the decision tree grows, the problem of sample size may arise. Hence, we use Tikhonov regularisation, axis parallel split regularisation null space regularisation for generating decision trees to full depth. In rotation based double random forest models, we used two transformations- principal component analysis and linear discriminant analysis, on randomly chosen feature subspace at each non-leaf node. Rotations on different random subspace features lead to more diverse decision tree ensembles and  better generalization performance.  
 Unlike standard random forest where the bootstrap aggregation is used at root node only, the proposed oblique and rotation double random forest use bootstrap aggregation at each non-terminal node for choosing the best split and then the original samples are sent down the decision trees. 
 The proposed double variants of the ensemble of decision trees results in bigger trees compared to the standard variants of the ensemble of decision trees.
  Experimental results and the statistical analysis  show the efficacy of the proposed oblique and rotation double random forest ensemble models  over standard baseline classifiers. Besides classification, we will expand this work to regression and times series forecasting problems in the future. Moreover, one can also perform benchmarking of the variants of the standard random forest, variants of double random forest and XGBoost to evaluate their performance on a common platform which can help in choosing the best model.  
\section*{Acknowledgment}
This work is supported by Science and Engineering Research Board (SERB), Government of India under Ramanujan Fellowship Scheme, Grant No. SB/S2/RJN-001/2016, and Department of Science and Technology under Interdisciplinary Cyber Physical Systems (ICPS) Scheme grant no. DST/ICPS/CPS-Individual/2018/276. We gratefully acknowledge the Indian Institute of Technology Indore for providing facilities and support.





\bibliographystyle{elsarticle-num-names}
\bibliography{refs.bib}







\end{document}